\documentclass{article}

\usepackage[preprint,nonatbib]{neurips_2024}

\usepackage[utf8]{inputenc} 
\usepackage[T1]{fontenc}    
\usepackage{hyperref}       
\usepackage{url}            
\usepackage{booktabs}       
\usepackage{amsfonts}       
\usepackage{nicefrac}       
\usepackage{microtype}      
\usepackage{xcolor}         
\usepackage{graphicx}
\usepackage{color}
\usepackage{threeparttable}
\usepackage{appendix}
\title{Conversational Disease Diagnosis via External Planner-Controlled Large Language Models}

\author{Zhoujian Sun$^1$, Cheng Luo$^1$, Ziyi Liu$^2$, Zhengxing Huang$^3$ \\ 
$^1$Zhejiang Lab, $^2$ Transtek Medical Electronic, $^3$ Zhejiang University \\
\texttt{sunzj@zhejianglab.com, zhengxinghuang@zju.edu.cn}\\
}

\begin{document}
\maketitle

\begin{abstract}
The development of large language models (LLMs) has brought unprecedented possibilities for artificial intelligence (AI) based medical diagnosis. However, the application perspective of LLMs in real diagnostic scenarios is still unclear because they are not adept at collecting patient data proactively. This study presents a LLM-based diagnostic system that enhances planning capabilities by emulating doctors. Our system involves two external planners to handle planning tasks. The first planner employs a reinforcement learning approach to formulate disease screening questions and conduct initial diagnoses. The second planner uses LLMs to parse medical guidelines and conduct differential diagnoses. By utilizing real patient electronic medical record data, we constructed simulated dialogues between virtual patients and doctors and evaluated the diagnostic abilities of our system. We demonstrated that our system obtained impressive performance in both disease screening and differential diagnoses tasks. This research represents a step towards more seamlessly integrating AI into clinical settings, potentially enhancing the accuracy and accessibility of medical diagnostics.
\end{abstract}

\section{Introduction}

Enabling artificial intelligence (AI) to diagnose disease has been a long-awaited goal since the concept of medical AI emerged \cite{yanase2019systematic}. The development of large language models (LLMs) brings unprecedented opportunities in AI-based diagnosis. Notably, Med-Palm 2 and GPT-4 Turbo have attained high scores on the United States Medical Licensing Examination  \cite{nori2023capabilities,singhal2023towards}. Recent research also illustrates that LLMs may perform as well as human doctors in many disease diagnostic tasks \cite{eriksen2023use,lee2023benefits,sandmann2024systematic,tu2024towards}.

Nonetheless, most LLMs are not adept at collecting patient data, which limits their application perspective. Almost all existing LLM-based studies formulate diagnosis as a question-answer task where LLMs are endowed with all necessary information to answer the diagnostic question \cite{eriksen2023use,lee2023benefits,sandmann2024systematic,saab2024capabilities,chen2023meditron}. In real diagnosis scenarios, doctors initially have no knowledge about the patient's condition, and patients also cannot comprehensively describe their own conditions. Distinguishing which information is useful and knowing when to collect the information are core skills of a doctor \cite{sokol2017listening}. If a LLM requires a doctor to collect all important information in advance to make a diagnosis, its practical value is quite doubtful, because the doctor usually already knows what disease the patient has when all information is collected. LLM-based diagnostic systems should be capable of collecting information from scratch and then proceeding to diagnosis. This demands that LLM-based diagnostic systems possess excellent planning abilities to \textbf{proactively ask dozens of appropriate questions} through interactions with patients. Most current LLMs lack such planning capabilities. For example, a recent study demonstrated that GPT-4 could achieve high diagnostic accuracy, ranging from 70\% to 90\%, when provided with complete patient information for diagnosing skin diseases. However, its accuracy can drop to 30\% to 60\% when it must diagnose starting from scratch \cite{johri2023testing}.

In this study, we aim to develop a LLM based diagnostic system that enhances planning capabilities by emulating doctors. Previous research suggests that medical consultations can roughly be divided into two distinct phases \cite{baerheim2001diagnostic}. In the first phase, which we call disease screening phase, doctors ask patients a series of questions mainly about their medical history and infer possible diseases based on the responses. This stage relies heavily on doctor's experience. In the second phase, which we call differential diagnosis phase, doctors ask questions to confirm or exclude the diseases suspected in the first phase. The questions asked during the differential diagnosis phase include the patient's laboratory test and medical examination results. This phase relies on objective medical knowledge. Due to the substantial differences between these two phases, we clearly need to develop two different planners when emulating doctors. The first should be data-driven, learning from extensive data on how to conduct consultations, while the second should be knowledge-driven, adhering strictly to medical knowledge and being interpretable.

We primarily face two challenges in implementing the two planners. (1) Real medical consultation dialogue datasets are scarce, which hampers the training of the first planner in a supervised manner. (2) Developing a decision procedure that adheres to medical literature typically requires expert involvement, making the second planner expensive and hard to maintain \cite{cowan2001expert}. In this study, we adopted a reinforcement learning (RL) approach to facilitate the autonomous training of the first planner without the need for expert demonstrations. We used LLMs to analyze patient admission records, identifying each symptom's presence or absence. Subsequently, we utilized a RL method to train the inquiry policy based on the structurized patient symptoms. Following this, we employed a neural network to predict high-risk diseases based on the outcomes of the inquiries. The decision procedure for diagnosing or ruling out diseases is implicitly recorded in medical literature in the form of natural text. Since leading LLMs have achieved capabilities nearly equivalent to junior doctors in natural language processing and medical knowledge, we attempt to summarize decision procedures by directly employing LLMs \cite{achiam2023gpt,singhal2023towards}. Additionally, we have designed a method that allows non-medical experts to refine these decision procedures, thereby reducing reliance on experts.

\begin{figure}[tb]
  \centering
  \includegraphics[width=11.12cm,height=3.7cm]{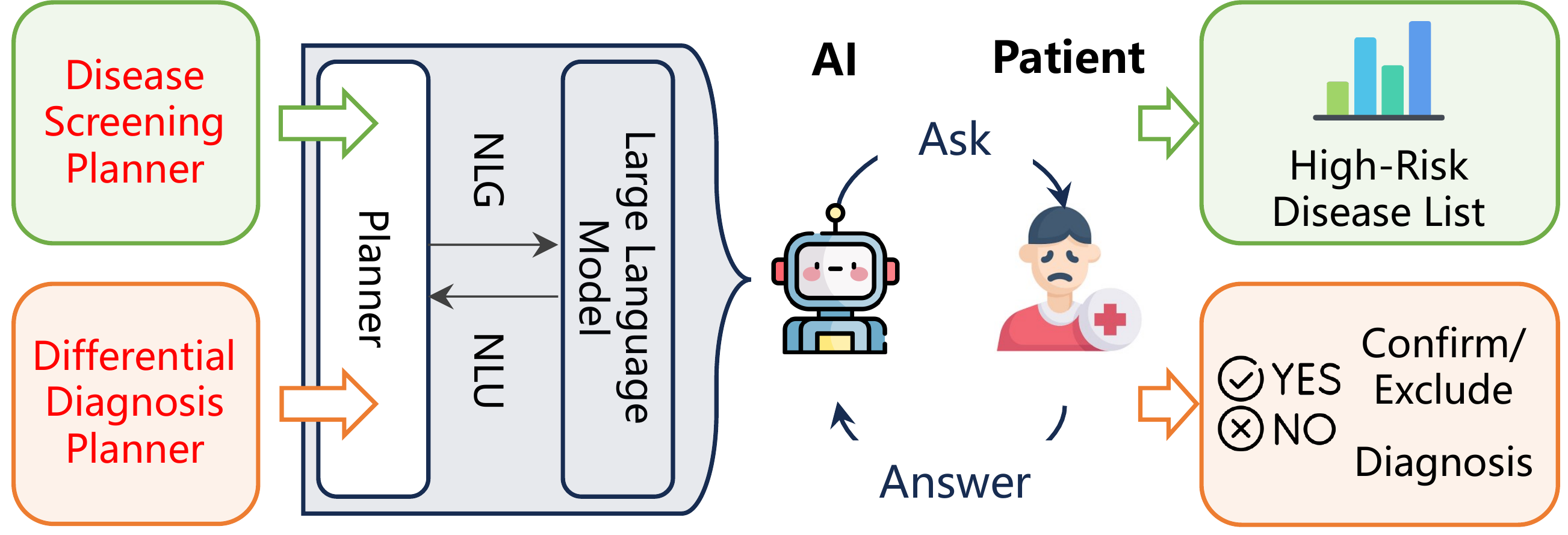}
  \caption{System Overview}
  \label{fig:modelFramework}
\end{figure}

We evaluated this study through retrospective simulated dialogues. We implemented a simulated doctor comprised of two planners and one LLM (Figure \ref{fig:modelFramework}). Planners are responsible for determining actions for each round, while the LLM handles the conversion of these actions into natural language and also parses responses into a format readable by the planners. We utilized another LLM to read real electronic medical record (EMR) data, simulating a patient who would respond to any questions posed by the simulated doctor. The simulated doctor was designed to ask a series of questions to diagnose the patient's illness. We conducted tests using the MIMIC-IV dataset \cite{johnson2023mimic}. The results show that our planners, controlling the LLM, achieved impressive performance in both phases. We contend that the proposed diagnostic system has the potential to enhance diagnostic precision and accessibility.
All source code and data are public available \footnote{https://github.com/DanielSun94/conversational\_diagnosis}.

\section{Related Work}
Recent research has illustrated the capabilities of LLMs in aiding disease diagnosis \cite{patel2023chatgpt,lee2023benefits,eriksen2023use,saab2024capabilities}. They use pre-collected patient information to conduct diagnosis. However, a LLM-based diagnostic system should prioritize and emphasize the capability to ask the right questions for information collection in multi-turn dialogue \cite{sokol2017listening}. Existing research seldom explores the multi-turn diagnostic capabilities of LLMs, and those that do usually only investigated LLMs' planning abilities in fewer than five rounds of free dialogue, which is far from sufficient to complete diagnostic tasks \cite{bao2023disc}. To our knowledge, the AMIE model is the only LLM trained to improve the medical information-gathering ability \cite{tu2024towards}. This model has two limitations. Firstly, it relies solely on patient self-reported symptoms for diagnosis, without incorporating data from medical laboratory tests or imaging reports. Since the symptoms of many diseases overlap, relying solely on symptoms for an accurate diagnosis is not only dangerous but also impractical. Secondly, the AMIE model is trained via a LLM-generated synthetic dialogue dataset. Without concrete evidence demonstrating that LLMs can match the efficacy of experts in medical consultations, the soundness of this approach is questionable.

Before the advent of LLMs, several diagnostic-oriented dialogue systems had been proposed \cite{wei2018task,lin2020towards,chen2023dxformer,chen2023benchmark}. However, their datasets are originated from online consultations, and the data quality was often questioned. In the past two years, research in various fields has explored ways to make LLMs proactively guide conversations and complete tasks. Most studies use prompt engineering, while some enhance LLMs' planning ability with external planners \cite{deng2023plug,hongru2023cue,fu2023improving}. However, these methods cannot be utilized in medicine because they still rely on public, high-quality, human annotated dialogue datasets, which in fact do not exist in the medical field. In this study, we will develop a dialogue system without using high-quality medical dialogue data.

LLM-based autonomous agent tackles complex tasks by decomposing them into simpler sub-tasks, addressing each one sequentially. Most research in this area utilizes prompt engineering tricks, e.g., chain of thought, to activate LLM's planning capabilities \cite{wei2022chain}. However, an autonomous agent typically does not require interaction with the user during the task completion process \cite{wang2024survey}. In contrast, our system needs to dynamically generate a plan based on the patient's responses. 

\begin{figure}[tb]
  \centering
  \includegraphics[width=13.11cm,height=9.04cm]{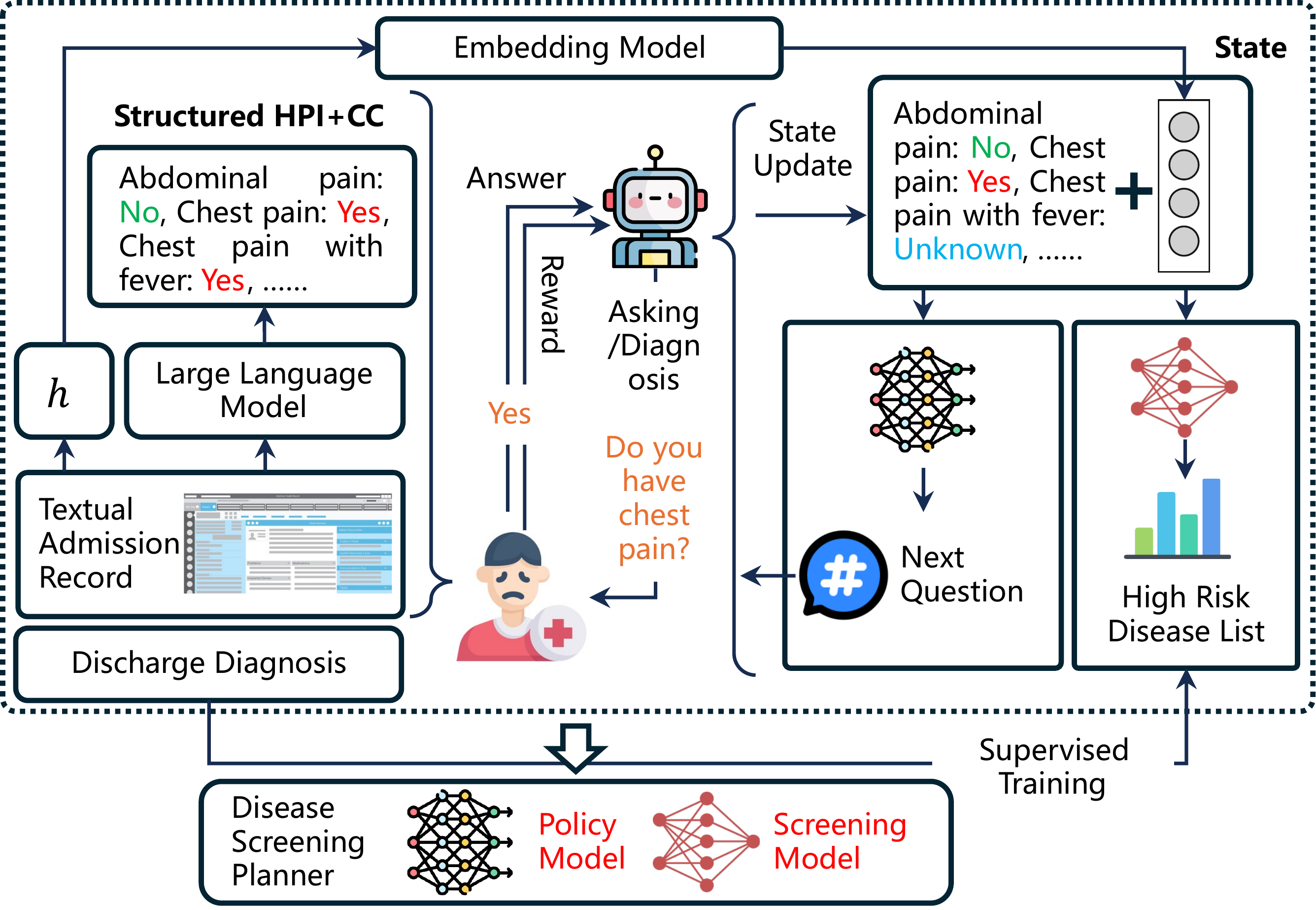}
  \caption{Diagnosis Screening Planner Optimizing}
  \label{fig:diseaseScreeningPlannerOptimizing}
\end{figure}

\section{Method}
\subsection{Disease Screening Planner}
Doctors are required to obtain important information from patients in medical consultations to formulate initial diagnoses, which includes their chief complaints (CC), history of present illness (HPI), past medical history, family history, social history, etc \cite{melms2021pilot}. HPI and CC are usually regarded as the most critical information, as they form the basis for at most 80\% of patient diagnoses \cite{hampton1975relative}. Despite its importance, collecting HPI and CC is challenging because patients usually cannot comprehensively report their symptoms \cite{sokol2017listening}. As patients may exhibit a subset of hundreds of possible symptoms, it is also impractical to exhaustively query every potential symptom. Doctors need to identify positive symptoms via their own experience with only several questions, thereby facilitating the formulation of a comparatively accurate diagnosis.

We employed an RL approach to train a policy model to ask questions and a supervised learning based screening model to conduct initial diagnosis (Figure \ref{fig:diseaseScreeningPlannerOptimizing}). We use $h$ to denote the past medical history, family history, and social history of a patient. As $h$ can typically be acquired through straightforward questioning, our study will not focus on collecting such information \cite{melms2021pilot}. We presume $h$ is textual and already known and its embedding is $e_h$. We use a planner to ask $N$ questions about patient symptoms in the RL paradigm.

\textbf{State Space}: 
The state $s_t=[e_h,p_t]$ is the concatenation of two elements. The first part is $e_h$, which is invariable in an episode. $p_t\in {\{0,1\}}^{3M}$ represents structured HPI and CC, where $M$ denotes the number of symptoms. Each symptom information is presented by a binary triplet, while $[0,0,1],[0,1,0],[1,0,0]$ means the symptom is unknown, confirmed, or denied, respectively. At the beginning, all symptoms are unknown, and their status is updated with every interaction.

\textbf{Action Space}: 
The action space contains $M$ ($N\ll M$) actions, where each question asks whether a related symptom is present in the next turn. We presume that there is a two-layer structure within the action space. The first layer refers to general symptoms (such as chest pain), while the secondary layer denotes more specific symptoms (such as chest pain accompanied by fever). Each second-layer symptom is affiliated with a first-layer symptom. We stipulate that the model can only inquire about second-layer symptoms after the patient acknowledges the affiliated first-layer symptoms. Meanwhile, we do not allow the planner to ask the same question twice.

\textbf{Reward}: 
We set the reward $R_t$ to one if the asked symptom exists, and to zero if the asked symptom is denied or not mentioned.

\textbf{Patient Agent}: 
We use admission records from patient EMRs to construct the patient agent. We first separated an admission record into $h$, HPI, and CC. Then, we structured the textual CC and HPI. Specifically, we utilized a LLM to determine whether patients had a symptom, and ultimately transformed the CC and HPI into a $M$-dimensional binary vector $p^{oracle}$. Each symptom is represented by one if it is confirmed, or zero if it is denied or not mentioned. A $h$ and a $p^{oracle}$ formulate a sample. When an agent receives a query from the planner, it can response the answer directly.

\textbf{Policy Model Learning}: 
We used an actor-critic model to generate the policy $\pi_t\in\mathbf{R}^M$, which is a stochastic vector, and the value $Q_t\in\mathbf{R}$ \cite{sutton2018reinforcement}. Each element in $\pi_t$ corresponds to an query action. The value of an element indicates the probability of the policy model selecting the corresponding action. We utilize a multi-layer perceptron (MLP) to learn a representation $r_t$ from $s_t$ and then use two MLPs to generate $\pi_t$ and $Q_t$ according to $r_t$, respectively. We adopted the proximal policy optimization (PPO) algorithm to train policy \cite{schulman2017proximal,stable-baselines3}. In this study, we preset a maximum number of inquiry rounds, and the PPO will train the RL agent to obtain the maximum reward within these rounds. To improve effectiveness, we assume that the patient agent will proactively disclose one first layer symptom to the RL agent before the first question. Given that patients typically start by describing their areas of discomfort in real medical consultations, we believe this design is justified.

\textbf{Screening Model Learning}:
After the policy model is optimized, we will use the final state of episodes to predict the initial diagnosis. As the patient discharge diagnosis is recorded in the EMR, we use a supervised learning method, i.e., MLP, to train the screening classifier.

\begin{figure}[tb]
  \centering
  \includegraphics[width=13.15cm,height=8.04cm]{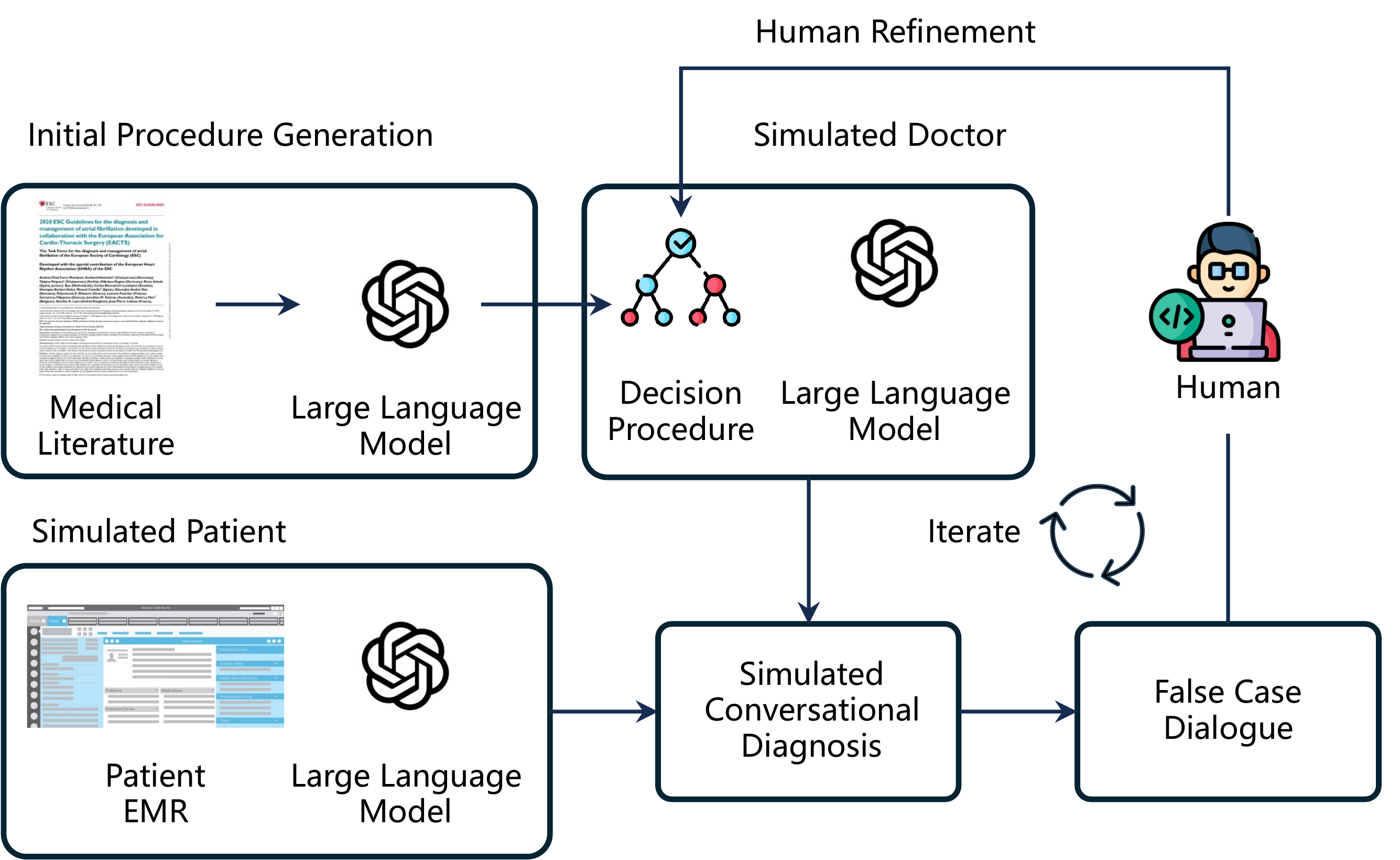}
  \caption{Differential Diagnosis Planner Optimizing}
  \label{fig:differentialDiagnosisPlannerOptimizing}
\end{figure}

\subsection{Differential Diagnosis Planner}
The differential diagnosis planner consists of a set of decision procedures, each corresponding to the diagnostic process for a specific disease. We use a LLM to transform medical literature and generate a preliminary structured decision procedure (Figure \ref{fig:differentialDiagnosisPlannerOptimizing}). The planner will parse the procedure and conduct inquiries according to the procedure. The effectiveness of the procedure is then tested in simulated diagnosis dialogues where one LLM with the external planner acts as a doctor and another as a patient, using complete discharge EMR for the patient role. In the simulated interaction, the doctor asks questions based on the decision procedure, and the patient responds based on the EMR contents. If the EMR does not contain the answer to a question, the patient simulator will indicate that the inquired physical condition is normal. Finally, the doctor simulator concludes with either a confirmed or excluded diagnosis. This outcome is then compared with the actual patient discharge diagnosis to identify any inaccuracies, creating a set of dialogues with incorrect conclusions. These results are analyzed to pinpoint which steps in the procedure led to incorrect diagnoses. According to the analyze result, the procedure undergoes refinement through revisions to its content. The refined decision procedure is then retested and improved iteratively. In this study, the refinement process was completed by a data scientist who does not hold a medical license.

\subsection{Simulated Dialogues}
We conducted retrospective simulated conversations between a doctor simulator and a patient simulator to evaluate the diagnostic performance of our system.

\textbf{Patient Simulator}:
We used a LLM to act as a patient. During the disease screening phase, we submitted a patient admission record to the LLM and instructed the LLM to answer questions based on the provided information (prompt is in appendix \ref{sec:patientSimulator}). This simulation method is widely used in previous studies \cite{tu2024towards,johri2023testing}. During the differential diagnosis phase, we submitted the entire EMR data, including laboratory tests and exam reports, as context to respond to the doctor simulator's inquiries.

\textbf{Doctor Simulator}:
We employed a LLM controlled by two planners to serve as a doctor in the disease screening phase and the differential diagnosis phase. The planners are responsible for generating questions or conducting diagnoses. The LLM is tasked with translating the questions generated by the two planners into plain language. It also interprets the patient's utterances and categorizes them into a format that the planners can process.

\section{Experiment}
\subsection{Dataset and Preprocessing}
We utilized textual clinical notes from the MIMIC-IV dataset to conduct experiments \cite{johnson2023mimic}. The MIMIC-IV contains EMRs for approximately 400,000 admissions to the intensive care unit (ICU) at Beth Israel Deaconess Medical Center in the USA between 2008 and 2019 (a sample is in appendix \ref{sec:emrSample}). We identified the 98 most common diseases from MIMIC-IV (happened larger than 200 times), excluding injuries from accidents and mental illnesses (disease list is in appendix \ref{sec:diseaseList}). We randomly selected 40,000 admissions with textual clinical notes from patients whose primary diagnosis (the first diagnosis in the discharge diagnoses) was among these 98 diseases. The reserved clinical notes contain 2,394 words in average, including admission records, discharge summaries, laboratory test results, imaging examinations reports, etc.

The symptom checker developed by the Mayo Clinic was utilized to structure HPI and CC \cite{mayo2024symptom}. This symptom checker categorizes common symptoms of patients into 28 first-layer categories and 689 second-layer symptoms. Each second layer symptom is associated with one of the first-layer categories. We used GPT-4 Turbo (1106 preview) to analyze the presence of these 717 symptoms (28+ 689) in each patient's HPI and CC information, thereby converting the textual admission EMR into a 717-dimensional binary vector (prompt in appendix \ref{sec:structurizeEMR}). We transformed the selected 40,000 clinical notes into binary vectors. After the transformation, we randomly selected 400 medical records for evaluation. The results show that GPT-4 Turbo's parsing performance is satisfactory, with both macro recall and macro precision are around 97\%, making it is suitable for constructing patient simulators for the development of the disease screening planner.

\subsection{Experimental Settings}
\textbf{Foundation Models:}
We used Microsoft Azure GPT-4 Turbo (04-09), GPT-4o (05-13), and Llama3 (70B-instruct) as foundation models of doctor simulators \cite{meta2024llama3,achiam2023gpt,OpenAI2024GPT4o}. We used GPT-4 Turbo (04-09) as the foundation model of patient simulators. GPT-4 Turbo and GPT-4o were chosen because they are two of the best LLMs. Llama3 was selected because it is a leading open-source LLM. Of note, transferring medical data outside hospitals often breaches data protection regulations, which means all closed LLMs (e.g., GPT-4o, Gemini-1.5 Pro , Claude-3 Opus) are actually unavailable in practice as they only operate through remote API \cite{reese2023limitations}. We need to ensure our system retains good diagnostic capabilities only with an open-source LLM which can be locally deployed.

\textbf{Disease Screening Settings:}
We divided the structured data of 40,000 patients' HPI and CC into three parts for training, validation, and testing, with proportions of 0.7, 0.1, and 0.2, respectively. All past medical history, social history, and family history information were directly provided, and we used a text embedding model to generate their embeddings \cite{openAI2024embedding}. We configured the system to have the patient initially report one positive first-layer symptom to the planner, who can then ask  additional 9 or 19 questions. The planner also trains a high-risk disease screening model based on the collected patient information and their primary discharge diagnosis. We used the planner to control a LLM to conduct 300 simulated conversations and evaluate performances.

\textbf{Differential Diagnosis Settings:}
We used heart failure as a case to test the planner's ability to make differential diagnoses. Heart failure is the most common disease in the MIMIC-IV dataset and one of the most complex common cardiovascular diseases \cite{groenewegen2020epidemiology}. We argue if our system can accurately diagnose heart failure, it may also be capable of diagnosing most other diseases. In this study, we used the heart failure diagnostic guideline published by the European society of cardiology as the source of diagnostic knowledge \cite{mcdonagh20212021}. We randomly selected 160 positive patients (i.e., primary diagnosis is heart failure) and 160 negative patients (i.e., heart failure is not in the discharge diagnosis) from the MIMIC-IV dataset. We randomly selected 120 records (with a 1:1 ratio of negative to positive) as the training set for the generation of the decision procedure and 200 records for testing the decision procedure. We used the GPT-4 Turbo (0125 preview) and the guideline to generate a preliminary decision procedure (prompt in appendix \ref{sec:decisionProceduregenerator}, result in appendix \ref{sec:prelinminaryDiagnosisProcedure}). Subsequently, we selected 40 samples from the training set for each round of simulated dialogue experiment. We summarized the incorrect diagnoses, had a non-expert review the causes of the errors, and modified the procedure based on the analysis results. This modification process was repeated three times. Finally, we conducted simulated diagnoses on the test data via the refined decision procedure (appendix \ref{sec:refinedDiagnosisProcedure}).

\textbf{Metrics:}
We evaluated performances through simulated dialogues between patients and doctors. During the disease screening phase, we assess the system using the Top-N hit rate of the real primary diagnosis in the ranking. For example, a Top 3 Hit rate of 0.50 means that in 50\% of test cases, the true primary discharge diagnosis appears within the top three predictions generated by the disease screening planner. In the differential diagnosis phase, the system can ask up to 20 questions. If the system provides a diagnosis within these 20 questions, the conversation is considered successful. Otherwise, it is deemed a failure, and the corresponding case is treated as a negative sample. We evaluated the differential diagnosis performance via accuracy, precision, recall, and F1 score.

\subsection{Disease Screening Performance}
\begin{table}[t]
\begin{threeparttable}
\caption{Disease Screening Performance}
\begin{tabular}{cccccccc}
\hline
\textbf{EP} & \textbf{\# Question} & \textbf{LLM} & \textbf{\# Sample} & \textbf{Top 1} & \textbf{Top 3} & \textbf{Top 5} & \textbf{Top 10} \\ \hline
No  & 10 & GPT-4 Turbo 04-09   & 300  & 0.273 & 0.510 & 0.597 & 0.717 \\ 
Yes & 10 & GPT-4 Turbo 04-09   & 300  & \textbf{0.330} & \textbf{0.550} & \textbf{0.637} & \textbf{0.770} \\
No  & 10 & Llama3-70B-Instruct & 300  & 0.240 & 0.423 & 0.483 & 0.583 \\ 
Yes & 10 & Llama3-70B-Instruct & 300  & 0.303 & 0.477 & 0.603 & 0.737 \\\hline
No  & 20 & GPT-4 Turbo 04-09   & 300  & 0.310 & 0.523 & 0.590 & 0.727 \\ 
Yes & 20 & GPT-4 Turbo 04-09   & 300  & 0.310 & 0.527 & 0.610 & 0.753 \\ 
No  & 20 & Llama3-70B-Instruct & 300  & 0.200 & 0.387 & 0.470 & 0.607 \\ 
Yes & 20 & Llama3-70B-Instruct & 300  & 0.317 & 0.493 & 0.603 & 0.747 \\ \hline
No  & 10 & GPT-4 Turbo 04-09   & 1000 & 0.279 & 0.507 & 0.597 & 0.714 \\
Yes & 10 & GPT-4 Turbo 04-09   & 1000 & 0.325 & 0.544 & 0.633 & 0.772 \\
\hline
\end{tabular}
\footnotesize{EP (external planner) set to ``yes'' means the LLM is controlled by an external planner and ``no'' means it operates independently, whose prompts are in appendix \ref{sec:screeningDoctorSimulator}. The number of questions is ten (20) because the external planner starts with one symptom and proactively asks nine (19) additional questions.}
\label{table:screeningPerformance}
\end{threeparttable}
\end{table}

Table \ref{table:screeningPerformance} provides a comparison of disease screening performance. We found that system's performance is similar when randomly selecting either 1000 or 300 cases. Thus, we argue that using 300 simulated dialogues is sufficient to evaluate our system's performance effectively. Incorporating an external planner (EP) significantly improves the system's disease screening capability. For example, without the EP, GPT-4 Turbo achieves a Top 1 Hit rate of 0.273, which increases to 0.330 under the guidance of the EP. In the absence of the EP, Llama3 significantly underperforms compared to GPT-4 Turbo, but the performance of Llama3 with the EP surpasses GPT-4 Turbo's without the EP. Meanwhile, we found simply asking more questions does not necessarily lead to better performance. For instance, when asking 20 questions, Llama3's performance significantly decreases, whereas GPT-4 Turbo only sees a significant improvement in its Top 1 Hit rate, with other metrics barely changing. This illustrates the inherent limitations in the planning capabilities of LLMs. Interestingly, even with the use of the EP, posing more questions does not yield performance improvements either. We delved deeper into the experimental results and investigated whether the performance improvement from the EP is due to its superior planning ability in the appendix \ref{sec:extendedScreeningAnalysis}. We also reported disease screening performance of the GPT-4o in the appendix \ref{sec:gpt4oScreeningPerformance}.

\subsection{Differential Performance}

\begin{table}[t]
\begin{threeparttable}
\caption{Differential Diagnosis Performance}
\begin{tabular}{cccccccc}
\hline
\textbf{EK} & \textbf{LLM} & \textbf{Success Rate} & \textbf{Accuracy}& \textbf{Precision}& \textbf{Recall}& \textbf{F1} \\\hline
None        & GPT-4 Turbo 04-09   & 86\%  & 86\% & 80\% & 95\%  & 87\% \\
Text        & GPT-4 Turbo 04-09   & 98\%  & 73\% & 65\% & \textbf{100\%} & 79\% \\
EP        & GPT-4 Turbo 04-09   & \textbf{100\%} & 82\% & 86\% & 76\%  & 80\% \\
EP+HF   & GPT-4 Turbo 04-09   & \textbf{100\%} & \textbf{91\%} & \textbf{89\%} & 92\%  & \textbf{91\%} \\ \hline
None        & Llama3-70B-Instruct & 0\%  & 50\% & 0\% & 0\%  & NA \\
Text        & Llama3-70B-Instruct & 15\%   & 63\%  & \textbf{96\%}  & 26\%   & 41\% \\
EP        & Llama3-70B-Instruct & \textbf{100\%} & 85\% & \textbf{96}\% & 73\%  & 83\% \\
EP+HF   & Llama3-70B-Instruct & \textbf{100\%} & \textbf{91}\% & 91\% & \textbf{90}\%  & \textbf{90\%} \\ \hline
\end{tabular}
\footnotesize{EK (external knowledge) column indicates the method of utilizing knowledge. ``None'' means no external knowledge is used; ``text'' uses diagnosis related text from the clinical guideline; ``EP'' employs a diagnostic procedure directly generated by the LLM into the external planner; ``EP+HF'' involves a human-corrected diagnostic procedure. Prompts of these experiments are recorded in appendix \ref{sec:diagnosisDoctorSimulator}.}
\label{table:differentialDiagnosisPerformance}
\end{threeparttable}
\end{table}

\begin{figure}[t]
  \centering
  \includegraphics[width=13.79cm,height=14.63cm]{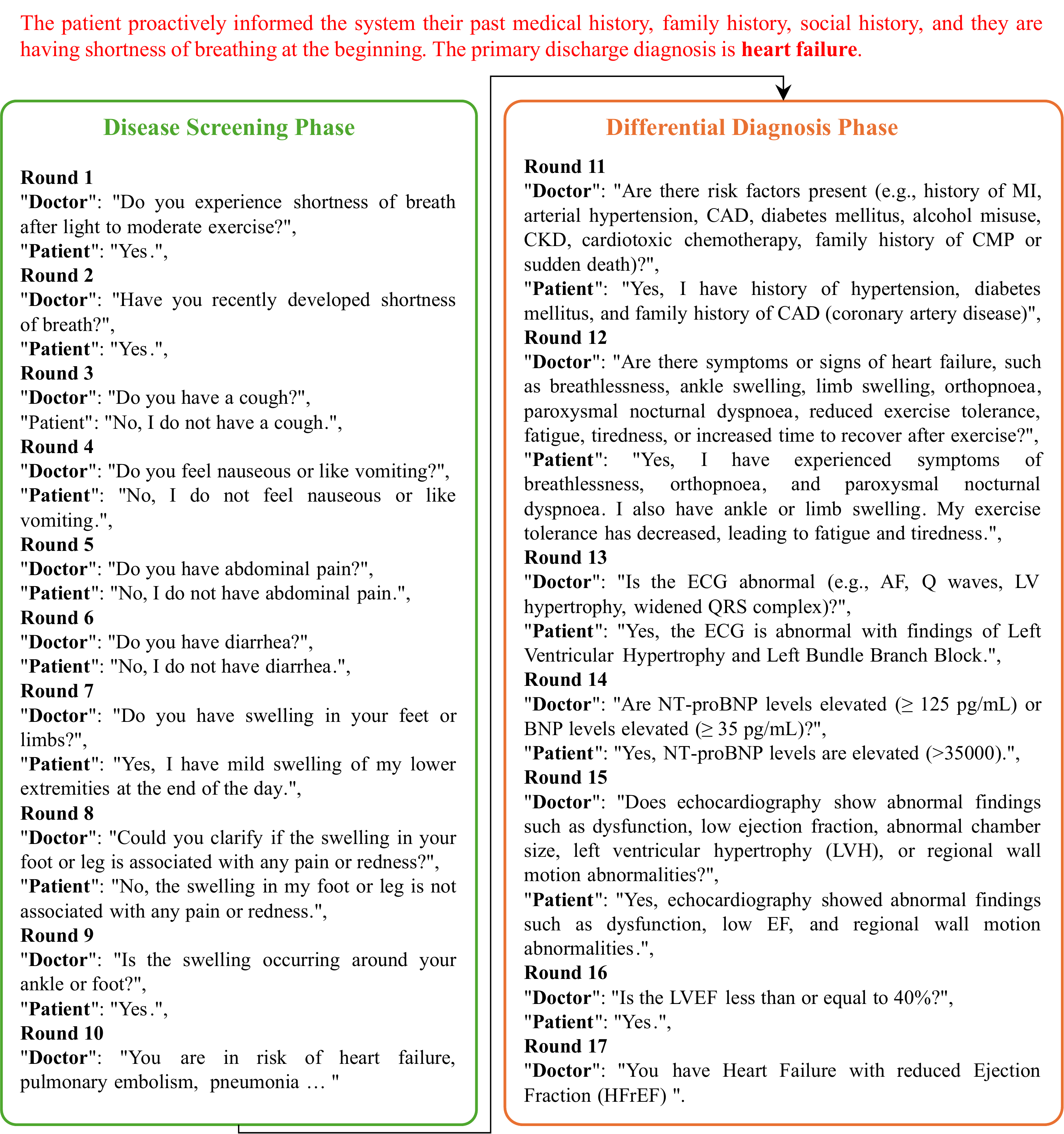}
  \caption{Conversational Diagnosis Sample}
  \label{fig:dialogueSample}
\end{figure}

We conducted differential diagnosis tests based on simulated dialogues using the decision procedure (results in Table \ref{table:differentialDiagnosisPerformance}). Llama3's success rate in dialogues is 0\% without external knowledge, indicating it doesn't know when to end a dialogue. Although GPT-4 Turbo significantly outperforms Llama3, it still experiences a 14\% failure rate in dialogues. By incorporating textual form knowledge, the performance of Llama3 significantly improved, with its F1 score increasing from NA to 41\%. The performance of GPT-4 Turbo actually decreased, indicating potential knowledge conflicts. Employing decision procedures has proven effective. However, the procedure directly extracted by the LLM is flawed, leading to low recall. This issue can be alleviated with non-expert human feedback (HF). The study demonstrates that the human refinement to the decision procedure improves the system's performance. GPT-4 Turbo’s success rate, accuracy and F1 scores can increase from 86\%, 86\% and 87\% to 100\%, 91\% and 91\%, respectively. The improvement is even more pronounced when using Llama3 as the foundation model, with the success rate soaring from 15\% to 100\%, and both accuracy and F1 scores escalating from 63\% and 41\% to 91\% and 90\%, respectively. This enhancement elevates the performance of open-source LLMs from unusable to fundamentally usable. We also reported differential diagnosis performance of GPT-4o in the appendix \ref{sec:gpt4oDifferentialPerformance}. 

In appendix \ref{sec:extendedDifferentialDiagnosis}, we conducted an error analysis to explore why our system still produces incorrect diagnoses in approximately 10\% of cases and more false negative errors compared to GPT-4 Turbo. The analysis shows that most of the false negative errors in our system originate from data quality issues, while most of the false positive errors stem from misdiagnosing patients in the pre-clinical phase. We investigated the performance of using a LLM to review dialogue results and then directly making diagnoses in the appendix as well.

\subsection{Dialogue Sample}

Figure \ref{fig:dialogueSample} illustrates a case in which the simulated patient initially informs the system (GPT-4 Turbo 04-09 as the foundation model) of their past medical history, family history, social history, and that they are experiencing difficulty breathing. The simulated doctor then conducted 17 dialogue rounds. According to the first ten rounds, the simulated doctor inferred that heart failure was a high-risk disease by asking about patient symptoms. Subsequently, the simulated doctor confirmed that the patient had heart failure by using the decision procedure extracted from the clinical guideline and refined by human. It is worth noting that since this study is retrospective, simulated patients possess all the information needed for a diagnosis. They just do not proactively disclose this information. In real prospective diagnostic scenarios, patients may not know the answers to the simulated doctor's questions either. Asking appropriate questions means the simulated doctor can issue appropriate prescriptions for the patients to undergo tests to obtain necessary information.

\section{Discussion and Conclusion}
This study introduces two planners to enhance LLMs' planning abilities. The disease screening planner improves diagnostic accuracy by utilizing RL to refine the inquiry policy. The differential diagnosis planner further enhances diagnoses by following evidence-based medical guidelines, translating these into structured decision procedures. Experimental results demonstrate that our system achieved impressive performance in conversational diagnostic tasks. To our knowledge, this is the first conversational diagnostic study conducted using real patient data rather than exam or question-answer datasets. Additionally, our system can be directly integrated into open-source LLMs, enabling them to handle dialogue diagnostic tasks comparable to closed LLMs.

Besides performance, our main contribution is the exploration of a new pipeline for utilizing EMRs in diagnostic dialogues. Existing relevant studies usually rely on synthetic dialogue data to fine-tune their LLMs and improving planning abilities, which is generated by larger LLMs such as Med-PaLM 2 or GPT-4 Turbo \cite{tu2024towards,chen2023huatuogptii}. However, this method is prohibitively expensive and unaffordable for most institutions. Our study results also suggest that even the most advanced LLMs struggle with conversational diagnostic tasks, casting doubt on the efficacy of synthetic data. Our study does not require dialogue data to fine-tune LLMs, which makes it cost-effective. By structuring textual HPI and CC via a LLM, our planners can be optimized through simulated experimentation, resulting in more effective inquiry policies than those of most current LLMs. Of note, our system surpasses GPT-4 Turbo by using only a standard symptom list and 40,000 EMRs. We argue that by carefully refining the symptom list and fully utilizing millions of EMRs already available in hospitals, the system's planning performance can be further enhanced.

The second contribution of this study is our improved handling of interpretability and reliability issues. The uninterpretability and hallucinations of LLMs raise concerns among doctors about their application perspective \cite{gilbert2023large}. We noticed that interpretability is not necessary during the disease screening phase, as it is often challenging for physicians themselves to clearly explain the rationale behind inquiries. We only need to ensure that LLM behavior adheres to medical literature in the differential diagnosis phase. We demonstrated that merely utilizing textual external knowledge may not improve LLM's planning capabilities to a satisfactory level. Thus, we explored a method that allows a LLM to autonomously generate decision procedures. The decision procedures can be presented in text form, understandable and verifiable by doctors, improving interpretability and reliability. Decision procedures also allow us to identify the causes of errors, which is challenging to accomplish through pure LLM-based diagnostics. Compared to traditional expert system research, we show that it is possible to generate decision procedures without the involvement of human experts \cite{cowan2001expert}. If disease diagnosis decision procedures can be auto-generated by LLMs with non-expert labor, it would enable rapid development of diagnostic workflows for each disease.

Although this study has achieved notable progress in conversational diagnosis, it is still a preliminary investigation with numerous limitations. The study relies on EMRs from ICU, involving patients with complex conditions. We also directly use an off-the-shelf RL algorithm. Consequently, the disease screening performance remains at a low level and requires enhancement. Due to budget limitations, this study only assessed differential diagnosis performance on heart failure. We plan to further refine the inquiry algorithm and seize opportunities to conduct clinical trials, testing the system's performance in real-world medical consultations.

\bibliography{reference}
\bibliographystyle{neurips}
\clearpage

\section*{Appendix}

\begin{appendices}

\section{Performance on GPT-4o} 
We also used GPT-4o as the foundation model of the doctor simulator \cite{OpenAI2024GPT4o}. However, due to the late release of GPT-4o and space limit, we only show experimental results of GPT-4o in the appendix.
\subsection{Screening Performance of GPT-4o} \label{sec:gpt4oScreeningPerformance}
\begin{table}[t]
\small
\centering
\caption{Disease Screening Performance of GPT-4o}
\begin{tabular}{cccccccc}
\hline
\textbf{EP} & \textbf{\# Question} & \textbf{LLM} & \textbf{\# Sample} & \textbf{Top 1} & \textbf{Top 3} & \textbf{Top 5} & \textbf{Top 10} \\ \hline
No & 10 & GPT-4o 05-13   & 300  & 0.297 & 0.537 & 0.610 & 0.733 \\
Yes & 10 & GPT-4o 05-13  & 300  & 0.320 & 0.533 & 0.613 & 0.747 \\
No & 20 & GPT-4o 05-13   & 300  & 0.327 & \textbf{0.553} & \textbf{0.653} & 0.747 \\
Yes & 20 & GPT-4o 05-13   & 300  & 0.310 & 0.540 & 0.637 & \textbf{0.777} \\
\hline
Yes & 10 & GPT-4 Turbo 04-09   & 300  & \textbf{0.330} & 0.550 & 0.637 & 0.770 \\
\hline
\end{tabular}
\label{table:screeningPerformanceGPT4o}
\end{table}

Table \ref{table:screeningPerformanceGPT4o} shows that GPT-4 Turbo (with the EP) obtained better performance than GPT-4o (with and without the EP) when asking ten questions, indicating utilizing our EP is beneficial. The performance of GPT-4o (with the EP) is also slightly better to GPT-4o (without the EP) when asking ten questions. However, when asking twenty questions, GPT-4o's performance significantly improved, surpassing GPT-4o (20 questions, with the EP) and comparable to GPT-4 Turbo (10 questions, with the EP). This result indicates that GPT-4o has significantly better planning capabilities than GPT-4 Turbo. However, these results do not directly indicate that GPT-4o has better planning capabilities than the EP. As shown in the Table \ref{table:RLAgentPerformanceVariousSettings}, when asked 20 questions, the performance of the EP (Top 1: 0.341, Top 3: 0.579, Top 5: 0.676, Top 10: 0.800) is better than that of GPT-4o. The decline in system performance observed in this subsection is due to errors in semantic parsing (appendix \ref{sec:semanticParsing}).

\subsection{Differential Performance of GPT-4o} \label{sec:gpt4oDifferentialPerformance}
\begin{table}[t]
\small
\centering
\caption{Differential Diagnosis Performance of GPT-4o}
\begin{tabular}{cccccccc}
\hline
\textbf{EK} & \textbf{LLM} & \textbf{Success Rate} & \textbf{Accuracy}& \textbf{Precision}& \textbf{Recall}& \textbf{F1} \\\hline
None        & GPT-4o 05-13  & 96\%  & 82\% & 74\% & 99\%  & 85\% \\
Text        & GPT-4o 05-13  & 86\%  & 85\% & 78\% & \textbf{100\%} & 87\% \\
None        & GPT-4 Turbo 04-09   & 86\%  & 86\% & 80\% & 95\%  & 87\% \\
Text        & GPT-4 Turbo 04-09   & 98\%  & 73\% & 65\% & \textbf{100\%} & 79\% \\
\hline
EP+HF       & GPT-4 Turbo 04-09   & \textbf{100\%} & \textbf{91\%} & \textbf{89\%} & 92\%  & \textbf{91\%} \\ \hline
\end{tabular}
\label{table:differentialDiagnosisPerformanceGPT4o}
\end{table}

Table \ref{table:differentialDiagnosisPerformanceGPT4o} indicates that the differential diagnosis performance of GPT-4o is comparable to GPT-4 Turbo, but both fall short of the performance of our EP-controlled GPT-4 Turbo. Without using any external knowledge, GPT-4o and GPT-4 Turbo achieve F1 scores of 85\% and 87\%, respectively. When external textual knowledge is introduced, the performance of GPT-4 Turbo declines, though its dialogue success rate improves. Conversely, GPT-4o sees a decrease in dialogue success rate but a slight improvement in F1. Both GPT-4o and GPT-4 Turbo tend to classify patients as positive cases, leading to low precision. The error analysis of this study (appendix \ref{sec:extendedDifferentialDiagnosis}) indicates that most of the system's errors are not due to the insufficiency of the foundation model. As utilizing the GPT-4o cannot avoid the errors, we did not conduct experiments on GPT-4o (with the EP).

\clearpage

\section{Extended Disease Screening Performance Analysis} \label{sec:extendedScreeningAnalysis}
\subsection{Reinforcement Learning Agent Performance}

\begin{table}[t]
\begin{threeparttable}
\centering
\small
\caption{Reinforcement Learning Agent Performance in Different Settings}
\begin{tabular}{lcccccccc}
\hline
\textbf{Method} & \textbf{\# Data} & \textbf{Reward} & \textbf{\# Que.} & \textbf{Top 1} & \textbf{Top 3} & \textbf{Top 5} & \textbf{Top 10}\\ 
\hline
PPO      &  100\%   & P   & 10 & 0.328$\pm$0.023 & 0.552$\pm$0.007 & 0.654$\pm$0.009 & 0.780$\pm$0.016 \\
PPO      &  100\%   & P   & 20 & 0.341$\pm$0.010 & 0.579$\pm$0.012 & 0.676$\pm$0.004 & 0.800$\pm$0.008 \\
PPO      &  100\%   & P   & 40 & 0.368$\pm$0.009 & 0.600$\pm$0.010 & 0.696$\pm$0.013 & 0.820$\pm$0.014 \\
\hline
PPO      &  100\%   & P+N & 10 & 0.317$\pm$0.017 & 0.554$\pm$0.029 & 0.655$\pm$0.014 & 0.777$\pm$0.011 \\
PPO      &  100\%   & P+N & 20 & 0.336$\pm$0.013 & 0.576$\pm$0.021 & 0.670$\pm$0.012 & 0.794$\pm$0.008 \\
PPO      &  100\%   & P+N & 40 & 0.363$\pm$0.011 & 0.597$\pm$0.013 & 0.703$\pm$0.009 & 0.818$\pm$0.005 \\
\hline
DQN      &  100\%   & P   & 10 & 0.326$\pm$0.021 & 0.547$\pm$0.009 & 0.647$\pm$0.016 & 0.779$\pm$0.011 \\
DQN      &  100\%   & P   & 20 & 0.337$\pm$0.013 & 0.571$\pm$0.012 & 0.669$\pm$0.014 & 0.795$\pm$0.010 \\
DQN      &  100\%   & P   & 40 & 0.360$\pm$0.011 & 0.591$\pm$0.017 & 0.694$\pm$0.009 & 0.815$\pm$0.012 \\
A2C      &  100\%   & P   & 10 & 0.316$\pm$0.027 & 0.554$\pm$0.014 & 0.653$\pm$0.013 & 0.666$\pm$0.015 \\
A2C      &  100\%   & P   & 20 & 0.337$\pm$0.016 & 0.577$\pm$0.012 & 0.672$\pm$0.011 & 0.793$\pm$0.011 \\
A2C      &  100\%   & P   & 40 & 0.358$\pm$0.012 & 0.594$\pm$0.009 & 0.691$\pm$0.008 & 0.812$\pm$0.012 \\
Rand     &  100\%   & P   & 10 & 0.246$\pm$0.014 & 0.419$\pm$0.012 & 0.518$\pm$0.016 & 0.666$\pm$0.008 \\
Rand     &  100\%   & P   & 20 & 0.251$\pm$0.007 & 0.433$\pm$0.013 & 0.535$\pm$0.008 & 0.678$\pm$0.009 \\
Rand     &  100\%   & P   & 40 & 0.262$\pm$0.007 & 0.452$\pm$0.006 & 0.556$\pm$0.007 & 0.692$\pm$0.011 \\
\hline
MLP       &  25\%   & P   & /  & 0.319$\pm$0.030 & 0.551$\pm$0.034 & 0.654$\pm$0.028 & 0.775$\pm$0.043 \\
MLP       &  50\%   & P   & /  & 0.358$\pm$0.027 & 0.598$\pm$0.035 & 0.702$\pm$0.019 & 0.825$\pm$0.031 \\
MLP       &  75\%   & P   & /  & 0.368$\pm$0.012 & 0.613$\pm$0.012 & 0.721$\pm$0.012 & 0.835$\pm$0.019 \\
MLP       &  100\%  & P   & /  & 0.373$\pm$0.010 & 0.620$\pm$0.011 & 0.723$\pm$0.009 & 0.842$\pm$0.013 \\
\hline
\end{tabular}
\footnotesize{\# Data means the fraction of data in the training dataset is used for experiment. Reward is P means the environment only give positive reward when its asked symptom is explicitly confirmed in the patient EMR. This reward strategy is used in the main text experiment. P+N means the environment give positive reward to the agent when its asked symptom is explicitly confirmed or denied. \# Que. means the number of questions.}
\label{table:RLAgentPerformanceVariousSettings}
\end{threeparttable}
\end{table}

We evaluated the performance of the screening RL agent (rather than system's performance in simulated dialogues) in different settings (Table \ref{table:RLAgentPerformanceVariousSettings}). Every experiment was repeated five times and we report the mean value and the standard error. The results show that the more questions asked, the better the performance of the agent. For instance, when asking 10 questions, the Top 3 hit rate of the experiment with PPO as the policy optimization method was 0.552, which increased to 0.600 when asking 40 questions.

In the main text, we only reward positive symptoms. In real medical consultations, symptoms denied by patients are also helpful for diagnosis. Therefore, we tested the model performance when rewards were given for denied symptoms as well (reward is set to P+N). Specifically, if the symptom asked about is one explicitly denied by the patient in the EMR, a reward of 0.2 was given. Experimental results show that the new reward strategy has almost no impact on model performance. On the other hand, we found that patient EMRs often do not record denied symptoms in detail. Therefore, we ultimately chose the setting in the main text that only rewards confirmed symptoms.

We conducted comparative experiments utilizing three methods: DQN, A2C, and a random inquiry policy \cite{stable-baselines3}. The results indicate that the performance of the random inquiry policy is significantly inferior to RL algorithms trained policies. This substantiates that the application of RL algorithms in this research significantly enhances the efficiency of information collection. Although the performance of the three RL algorithms is comparable, the planner trained using PPO slightly outperforms those trained with DQN and A2C in terms of diagnostic capabilities.

We also examined the impact of the size of training data on model performance. This experiment used all historical data, structured HPI and CC, and conducted predictions via the MLP. Generally, the larger the training dataset, the better the screening performance of the model, but the marginal gains decrease as the dataset size increases. The results demonstrate that the model performance using 75\% of training samples was very similar to that obtained with the full training dataset. This is also why the study only included 40,000 patient admission records.

We also attempted to fuse the policy model and the screening model, setting disease diagnosis as a type of action in the experimental setup. The algorithm failed to converge. Therefore, we have not reported the experimental results under this setting.

\subsection{Semantic Parsing Consistency Analysis} \label{sec:semanticParsing}
\begin{table}[t]
\small
\centering
\caption{Symptom Consistency Analysis}
\begin{tabular}{lccccccc}
\hline
\textbf{LLM} & \textbf{EP} & \textbf{\# Question} & \textbf{Accuracy} & \textbf{Precision} & \textbf{Recall} & \textbf{F1} \\ 
\hline
GPT-4 Turbo 04-09   &  Yes  & 10 & 83\% & 73\% & 90\% & 78\% \\ 
GPT-4 Turbo 04-09   &  Yes  & 20 & 82\% & 62\% & 88\% & 71\% \\ 
GPT-4o 05-13        &  Yes  & 10 & 80\% & 70\% & 83\% & 73\% \\ 
GPT-4o 05-13        &  Yes  & 20 & 81\% & 62\% & 82\% & 68\% \\ 
Llama3-70B-Instruct &  Yes  & 10 & 82\% & 71\% & 88\% & 76\% \\ 
Llama3-70B-Instruct &  Yes  & 20 & 81\% & 60\% & 86\% & 68\% \\ 
\hline
\end{tabular}
\label{table:screeningErrorAnalysis}
\end{table}

The performance of the RL agent differs from that in simulated dialogues. As the number of questions posed by the RL agent increases, its performance gradually improves. However, in simulated dialogues, when asking 20 questions, the performance of the LLM based system did not improve significantly. In fact, it might even have declined. This discrepancy may be attributed to the information distortion in dialogues. In simulated dialogues, the doctor simulator must convert the RL agent's actions into natural language outputs. The patient simulator then interprets these questions using the patient's EMR and responds in natural language. Finally, the doctor simulator understands the patient's response. This sequence includes two times of natural language understanding and two times of natural language generation. We conducted analyses on the consistency between the simulated dialogue generated state and the original structured states (Table \ref{table:screeningErrorAnalysis}). We found that all GPT-4 Turbo, GPT-4o, and Llama3-70B-instruct tend to judge symptoms identified as negative in the structured HPI and CC as positive in dialogues. The consistency of GPT-4o is comparable to Llama3, and the consistency of Llama3 is lower than GPT-4 Turbo. More critically, our experiments show that as the RL agent asks more questions, it is more likely to extracts symptom that are inconsistent with the structured HPI and CC. This leads to a substantial divergence between the state distribution received by the RL during simulated dialogue and the state distribution used during training, causing performance degradation when we ask more questions in simulated dialogues. In the future work, we will attempt to address the issue of semantic parsing errors.

\subsection{Disease Disease Screening Performance with Different Data}

\begin{table}[t]
\small
\caption{Disease Screening Performance with Different Data}
\centering
\begin{tabular}{lccccccc}
\hline
\textbf{Data Type} & Method&\textbf{Top 1} & \textbf{Top 3} & \textbf{Top 5} & \textbf{Top 10} \\ \hline
PH                     & Embedding+MLP & 0.231 & 0.409 & 0.512 & 0.653 \\
Structured HPI+CC      & MLP & 0.324 & 0.560 & 0.663 & 0.787 \\
PH + Structured HPI+CC & Embedding+MLP & 0.373 & 0.620 & 0.723 & 0.842 \\
Full Data              & LLM Zero Shot (300 cases)& 0.433 & 0.677 & 0.733 & 0.823 \\
Full Data              & Embedding+MLP & 0.611 & 0.838 & 0.904 & 0.956 \\
\hline
\end{tabular}
\label{table:extendedScreeningPerformance}
\end{table}

\begin{figure}[htbp]
  \centering
  \includegraphics[width=13.9cm,height=6.24cm]{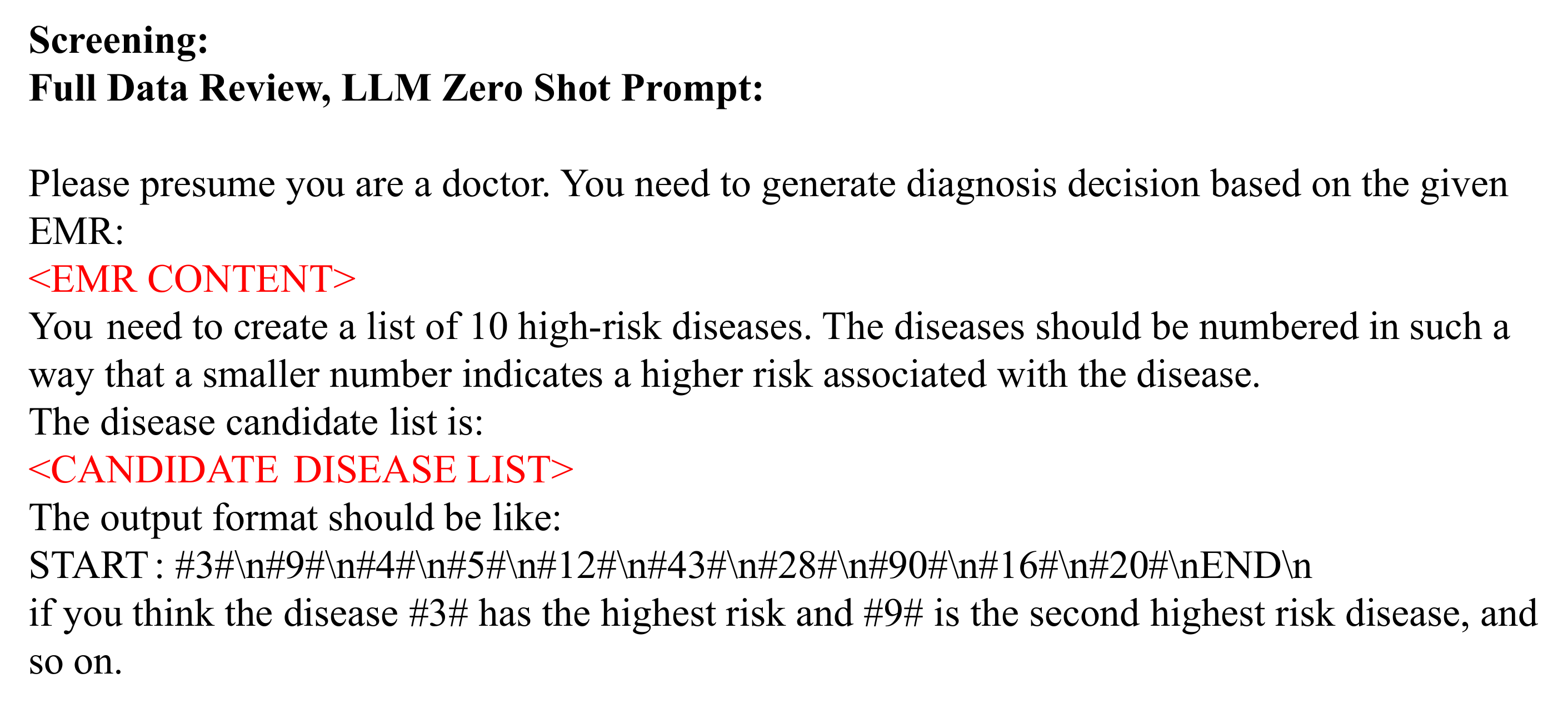}
  \caption{Full Data Review, LLM Zero Shot Prompt (Screen)}
  \label{fig:fullDataReviewZeroShot}
\end{figure}

We tested the disease screening performance in five different scenarios (Table \ref{table:extendedScreeningPerformance}). (1) PH, where we first used a text embedding model to generate a representation of past medical history, social history, and family history, followed by disease screening through a MLP. (2) Structured HPI+CC, where we used structured history of present illness and chief complaint for disease screening through a MLP. (3) PH + Structured HPI+CC, where the representations of PH and Structured HPI+CC were concatenated, and disease screening was conducted through a MLP. (4) Using complete textual admission records, and then employing GPT-4 Turbo for disease screening in a zero-shot learning manner (prompt in Figure \ref{fig:fullDataReviewZeroShot}). (5) Using a text embedding model to generate embeddings for the complete admission data, followed by disease screening through a MLP \cite{openAI2024embedding}.

The results show that both PH and structured HPI+CC are effective for disease screening, and their combined use outperforms the use of either information alone. If GPT-4 Turbo directly reads the complete patient information, its Top 1 Hit rate is about 0.433. In the experiment from main text, when GPT-4 Turbo conducted simulated dialogues, its Top 1 Hit rate was about 0.310, showing a significant performance difference. This suggests that the ability of LLMs to diagnose from scratch is significantly inferior to diagnose with all patient information. This phenomenon was also found in other studies \cite{johri2023testing}. We also tested the upper limit of disease screening performance. In the fifth scenario, the model's Top 1 Hit rate reached 0.611, whereas our system's Top 1 Hit rate in the experiment was only 0.330. This result indicates that there is considerable room for performance enhancement in the implementation of the disease screening planner.

\subsection{Symptom Collecting Efficacy Analysis}
\begin{table}[t]
\small
\caption{Symptom Collecting Efficacy Analysis Result}
\centering
\begin{tabular}{lccccc}
\hline
\textbf{EP} &\textbf{Top 1} & \textbf{Top 3} & \textbf{Top 5} & \textbf{Top 10} \\ \hline
Yes                  & 0.144$\pm$0.010 & 0.256$\pm$0.013 & 0.308$\pm$0.008 & 0.420$\pm$0.012 \\
No                   & 0.124$\pm$0.008 & 0.260$\pm$0.011 & 0.296$\pm$0.009 & 0.356$\pm$0.013 \\
\hline
\end{tabular}
\label{table:symptomCollectingAnalysis}
\end{table}
This subsection aims to answer a question: Does the improved performance of our screening planner arise from superior inquiry policies compared to GPT-4 Turbo, or from the application of supervised learning-based diagnoses? This problem presents substantial challenges for direct exploration. Thus, we investigated it indirectly. We produced 1,000 inquiry dialogues via GPT-4 Turbo and another 1,000 where the planner controlled GPT-4 Turbo. We then utilized a text embedding model to create representations of these dialogues, as well as corresponding past medical, social, and family histories. Following this, we applied supervised learning techniques to predict diseases based on these text representations. Our hypothesis posits that representations derived from more effective inquiry policies should achieve better classification outcomes.

We divided the dialogue dataset into three parts according to the proportions of 0.7, 0.1, and 0.2 for training, validation, and testing, respectively. Every experiment was repeated five times and we report the mean value and the standard error (Table \ref{table:symptomCollectingAnalysis}). The experimental findings suggest that the inquiry policies implemented by the EP obtained slightly better performance. This result indirectly supports the possibility that the policies developed by an RL-based planner could be superior to those based on GPT-4 Turbo.

\clearpage

\section{Extended Differential Diagnosis Analysis} \label{sec:extendedDifferentialDiagnosis}

\subsection{Error Analysis} 
\begin{table}[t]
\small
\caption{Differential Diagnosis Error Analysis}
\centering
\begin{tabular}{lcp{8.5cm}}
\hline
\textbf{Patient Identifier} & \textbf{Error} & \textbf{Reason} \\ 
\hline
11877234-23511109  &  FN  & He did not manifest any HF symptoms  \\ 
12882274-25494221  &  FN  & Echocardiogram report lost\\ 
13207228-26473644  &  FN  & Patient used a not typical HF symptom to describe its feeling.  \\ 
13017386-20509990  &  FN  & BNP result lost  \\ 
16402803-20664897  &  FN  & BNP result lost  \\ 
13605998-29618026  &  FN  & BNP result lost  \\ 
17536222-26218111  &  FN  & BNP result lost \\ 
11842519-24716312  &  FN  & Echocardiogram report lost  \\ 
11059748-25135118  &  FP  & Preclinical HF, symptoms induced by myocardial infarction \\
13695790-23923538  &  FP  & Preclinical HF, symptoms induced by coronary artery disease \\
14744884-25708947  &  FP  & Preclinical HF, symptoms induced by postural hypotension \\
14839161-23286377  &  FP  & Preclinical HF, symptoms induced by respiratory diseases \\
16316762-26322596  &  FP  & Preclinical HF, symptoms induced by respiratory diseases \\
15629520-27947176  &  FP  & Preclinical HF, symptoms induced by myocardial infarction \\
16609565-22342923  &  FP  & Preclinical HF, symptoms induced by bacterial peritonitis \\
17079194-25621293  &  FP  & Preclinical HF, symptoms induced by reactive airway disease \\
17232630-29523740  &  FP  & Preclinical HF, symptoms induced by respiratory failure \\
19241989-27097138  &  FP  & LLM error \\
19826828-24884026  &  FP  & Preclinical HF, symptoms induced by pneumonia \\
\hline
\end{tabular}
\label{table:DifferentialDiagnosisErrorAnalysis}
\end{table}

We investigated why false positives (FP) and false negatives (FN) still occur during the disease differential diagnosis phase, even when diagnoses are conducted according to the guideline. We analyzed simulated dialogues generated by GPT-4 Turbo (Table \ref{table:DifferentialDiagnosisErrorAnalysis}) and identified eight false negative cases. Six of these were due to incomplete data. In the simulated dialogues, we assumed that if an answer to a question was not recorded in the EMR, the patient would respond that the asked question is normal. If important information was missing, the doctor simulator would likely be misled. Therefore, the system's lower recall compared to GPT-4 Turbo and GPT-4o should not be simply interpreted as the model missing diagnoses. If the data in these six cases had been complete, the errors might not have occurred, and the recall of our system will be comparable to GPT-4 Turbo and GPT-4o. Among the remaining two cases, one involved a patient admitted before symptoms manifested, with no signs of heart failure, which led directly to a diagnostic error. Only one case resulted from the inadequacies of our planner itself. The patient described their symptoms with the term ``chest weakness,'' an atypical term for shortness of breath which is not recorded in the procedure, resulting in an error.

We identified 11 false positive cases. In ten of these, the reasons for the false positives were similar. These patients already had structural cardiac abnormalities, and many had a history of myocardial infarction. When such patients exhibited typical symptoms of heart failure, like chest discomfort or ankle swelling due to other diseases, they passed through the decision procedure and were considered positive cases, although they did not have a discharge diagnosis of heart failure. They should be considered as being in a ``pre-clinical phase of heart failure'' \cite{young2021progression}. Only one false positive case was due to a parsing error by the LLM. 

In summary, we argue that our system is generally reliable, while most errors are not caused by issues with the decision procedures generated in this study.

\subsection{Dialogue Review based Diagnosis Analysis}
\begin{table}[t]
\small
\caption{Differential Diagnosis by Reviewing Dialogue}
\centering
\begin{tabular}{cccccccc}
\hline
\textbf{E. K.}& \textbf{LLM} & \textbf{Knowledge} & \textbf{\# Question} & \textbf{Accuracy}&\textbf{Precision}&\textbf{Recall}&\textbf{F1}\\ 
\hline
None & GPT-4 Turbo 04-09 & No  & 8   & 77\%  & 69\%  & 100\%  & 81\%  \\
None & GPT-4 Turbo 04-09 & No  & 20  & 73\%  & 65\%  & 100\%  & 79\%  \\
None & GPT-4 Turbo 04-09 & Yes & 8   & 78\%  & 70\%  & 100\%  & 82\%  \\
None & GPT-4 Turbo 04-09 & Yes & 20  & 77\%  & 69\%  & 100\%  & 82\%  \\
Text & GPT-4 Turbo 04-09 & No  & 8   & 79\%  & 73\%  & 94\%  & 82\%  \\
Text & GPT-4 Turbo 04-09 & No  & 20  & 79\%  & 71\%  & 99\%  & 83\%  \\
Text & GPT-4 Turbo 04-09 & Yes & 8   & 81\%  & 75\%  & 92\%  & 83\%  \\
Text & GPT-4 Turbo 04-09 & Yes & 20  & 82\%  & 74\%  & 99\%  & 85\%  \\
\hline
None & GPT-4o 05-13& No  & 8   & 84\%  & 78\%  & 95\%  & 86\%  \\
None & GPT-4o 05-13& No  & 20  & 84\%  & 76\%  & 99\%  & 86\%  \\
None & GPT-4o 05-13& Yes & 8   & 86\%  & 80\%  & 94\%  & 87\%  \\
None & GPT-4o 05-13& Yes & 20  & 85\%  & 77\%  & 99\%  & 87\%  \\
Text & GPT-4o 05-13& No  & 8   & 84\%  & 76\%  & 100\%  & 87\%  \\
Text & GPT-4o 05-13& No  & 20  & 79\%  & 71\%  & 100\%  & 83\%  \\
Text & GPT-4o 05-13& Yes & 8   & 85\%  & 79\%  & 97\%  & 87\%  \\
Text & GPT-4o 05-13& Yes & 20  & 83\%  & 75\%  & 99\%  & 85\%  \\
\hline
None & Llama3-70B-Instruct  & No  & 8   & 73\%  & 65\%  & 98\%  & 78\%  \\
None & Llama3-70B-Instruct        & No  & 20  & 72\%  & 64\%  & 100\%  & 78\%  \\
None & Llama3-70B-Instruct        & Yes & 8   & 75\%  & 66\%  & 98\%  & 79\%  \\
None & Llama3-70B-Instruct        & Yes & 20  & 71\%  & 62\%  & 100\%  & 77\%  \\
Text & Llama3-70B-Instruct        & No  & 8   & 78\%  & 70\%  & 99\%  & 82\%  \\
Text & Llama3-70B-Instruct        & No  & 20  & 75\%  & 66\%  & 100\%  & 80\%  \\
Text & Llama3-70B-Instruct        & Yes & 8   & 79\%  & 71\%  & 99\%  & 83\%  \\
Text & Llama3-70B-Instruct        & Yes & 20  & 75\%  & 67\%  & 99\%  & 80\%  \\
\hline
\end{tabular}
\label{table:DifferentialDiagnosisReviewAnalysis}
\end{table}

When the LLM is not controlled by an EP, it may not be able to complete the diagnosis task within 20 rounds of dialogue. Here, we use a LLM to diagnose heart failure by reviewing corresponding dialogues (Table \ref{table:DifferentialDiagnosisReviewAnalysis}). We evaluated two scenarios: one using the first eight rounds (as the EP asks up to eight questions) as the context and the other using the all 20 rounds of dialogues. We also tested the performance with and without using diagnostic knowledge texts to help diagnosis (Knowledge column set to Yes or No). Prompts are in appendix \ref{sec:dialogueReview}. The results show that when using a LLM to review dialogues for diagnosis, there is little difference in performances between using eight rounds of dialogue and 20 rounds of dialogue. At the same time, the effect of using diagnostic knowledge text is also insignificant. The diagnostic performance based on LLM reviewing is significantly lower than the diagnostic performance based on our system (accuracy 91\%, F1 91\%).

\clearpage

\section{Heart Failure Diagnosis Procedure} \label{sec:diagnosisProcedure}
\subsection{Preliminary Heart Failure Diagnosis Procedure} \label{sec:prelinminaryDiagnosisProcedure}
\begin{figure}[h]
  \centering
  \includegraphics[width=13.9cm,height=17.29cm]{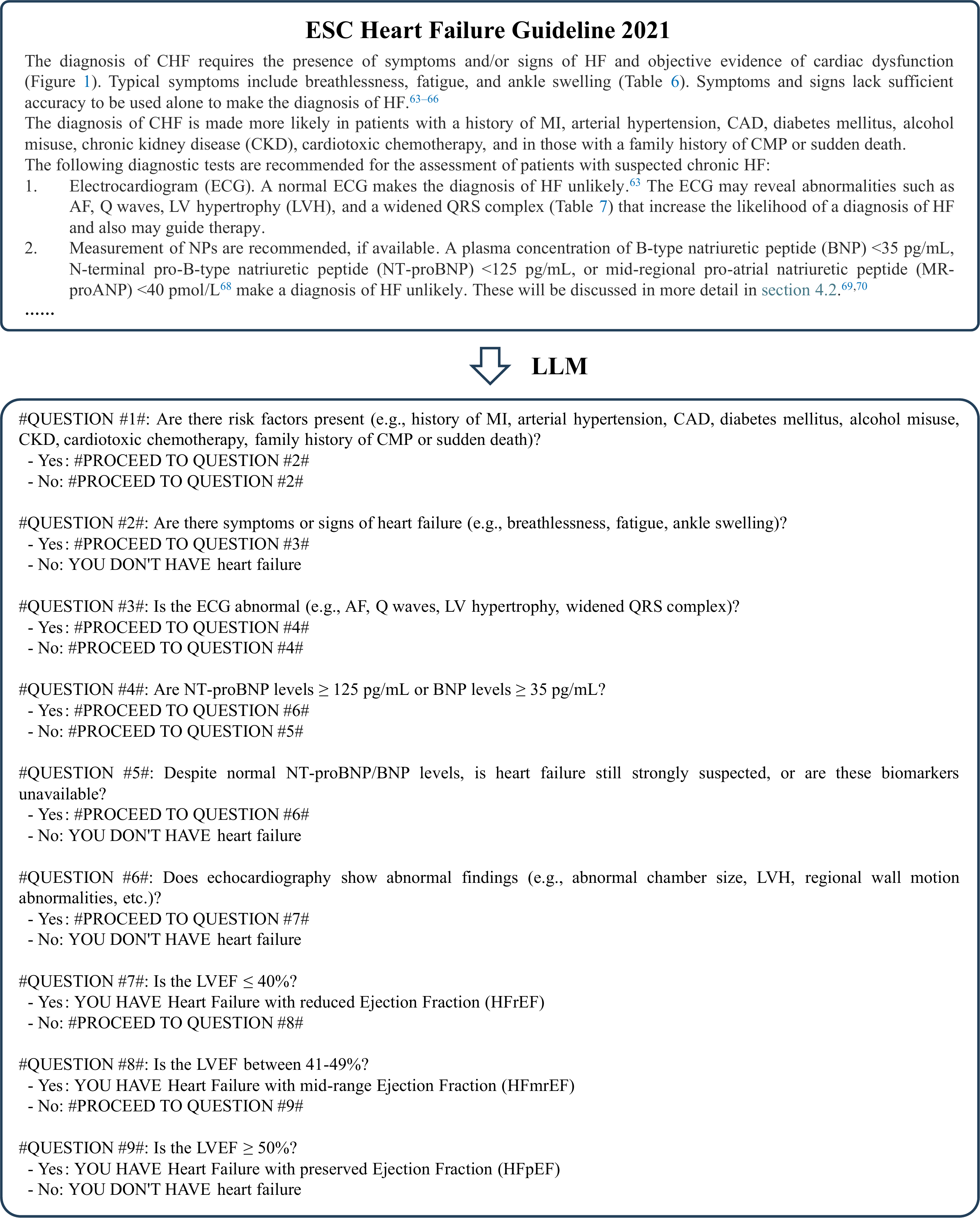}
  \caption{Decision Procedure Directly Generated By the GPT-4 Turbo}
  \label{fig:prelinminaryDiagnosisProcedure}
\end{figure}
Figure \ref{fig:prelinminaryDiagnosisProcedure} shows the decision procedure generated by the GPT-4 Turbo. As decision procedure is understandable and primarily based on the clinical guideline, it may resolve doctors' concerns about the reliability of AI.  
\clearpage

\subsection{Refined Heart Failure Diagnosis Procedure} \label{sec:refinedDiagnosisProcedure}
\begin{figure}[h]
  \centering
  \includegraphics[width=13.9cm,height=15.48cm]{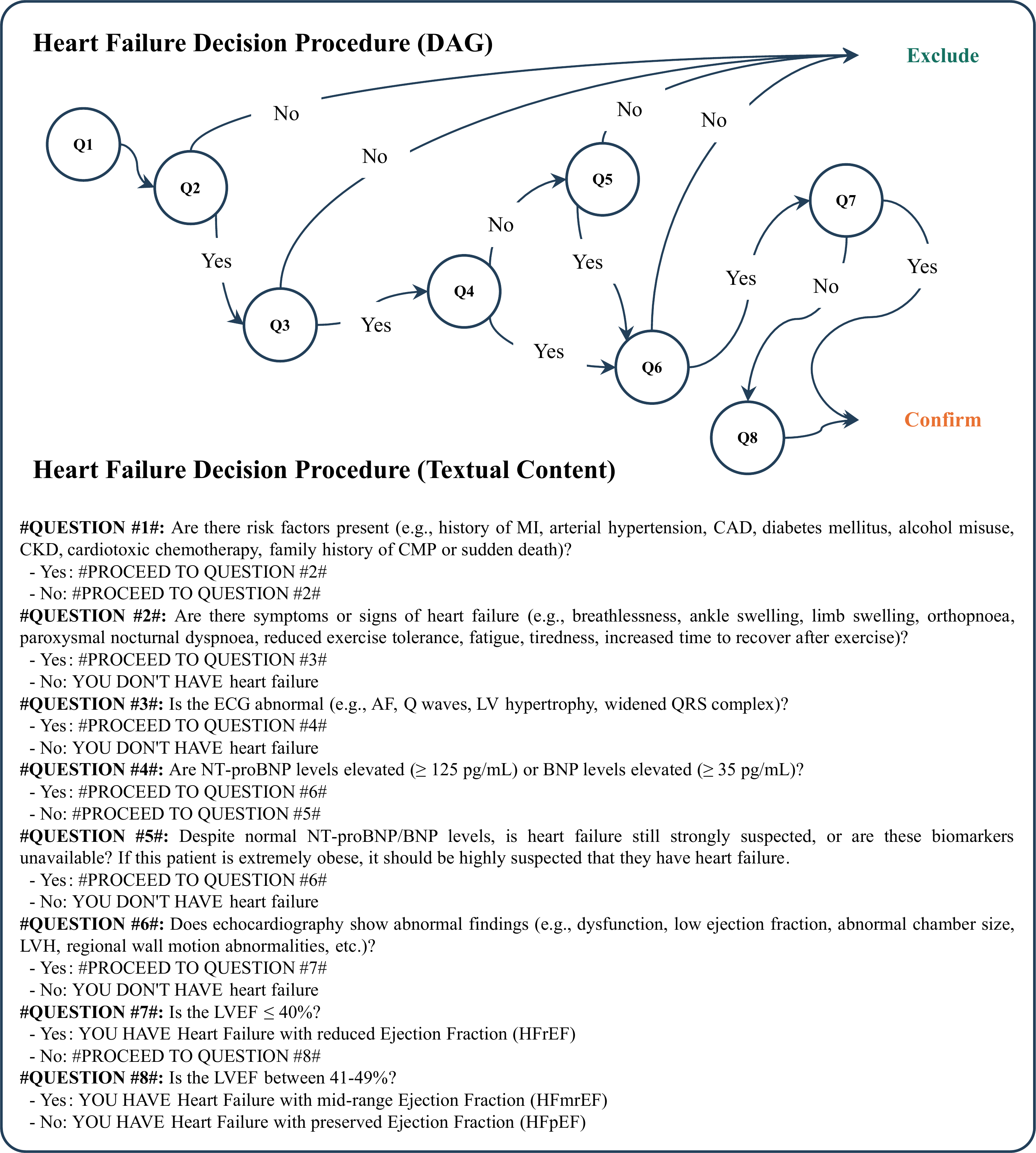}
  \caption{Decision Procedure after Human Refinement}
  \label{fig:refinedDiagnosisProcedure}
\end{figure}
Figure \ref{fig:refinedDiagnosisProcedure} shows the decision procedure optimized through the human feedback process. The decision procedure is a directed acyclic graph with eight nodes. There is a unique starting node, and each node corresponds to a question that needs to be asked by the doctor, with each question answerable by ``yes'' or ``no''. Ultimately, this graph can output ``confirm'' or ``exclude''.  It is evident that the diagnostic procedure generated directly by the LLM is of high quality. The human refinement only involved adding a few symptoms that were omitted in the original procedure, without significantly altering the content and structure. The effectiveness of the diagnostic procedure extracted by non-experts and the LLM has also been advocated by our clinical collaborators.
\clearpage

\section{Prompts}
\subsection{Admission Structurize Prompt} \label{sec:structurizeEMR}
\begin{figure}[htbp]
  \centering
  \includegraphics[width=13.9cm,height=9.66cm]{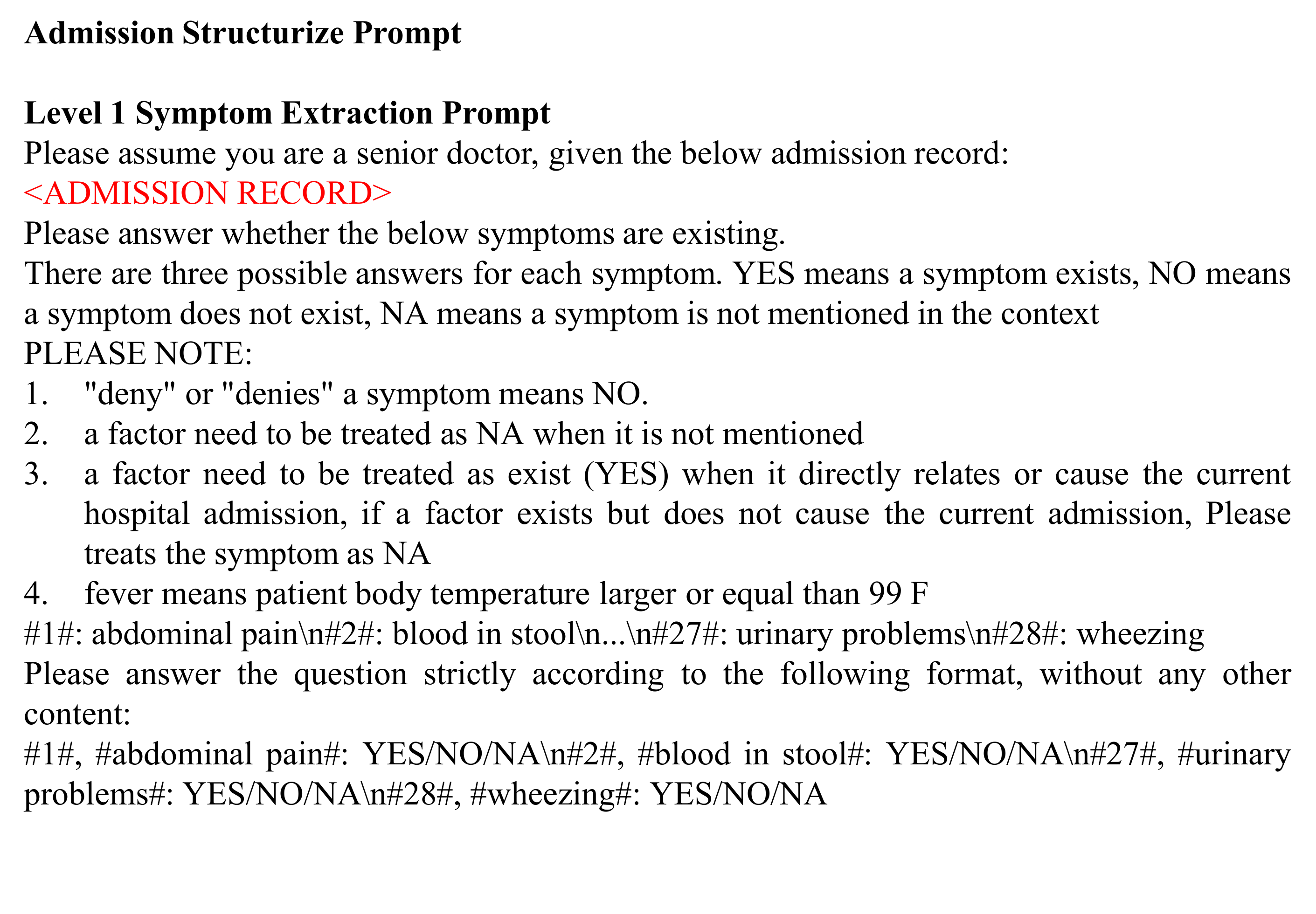}
  \caption{Admission Record Structurize Prompt 1}
  \label{fig:admissionStructurize1}
\end{figure}

\begin{figure}[htbp]
  \centering
  \includegraphics[width=13.9cm,height=11.37cm]{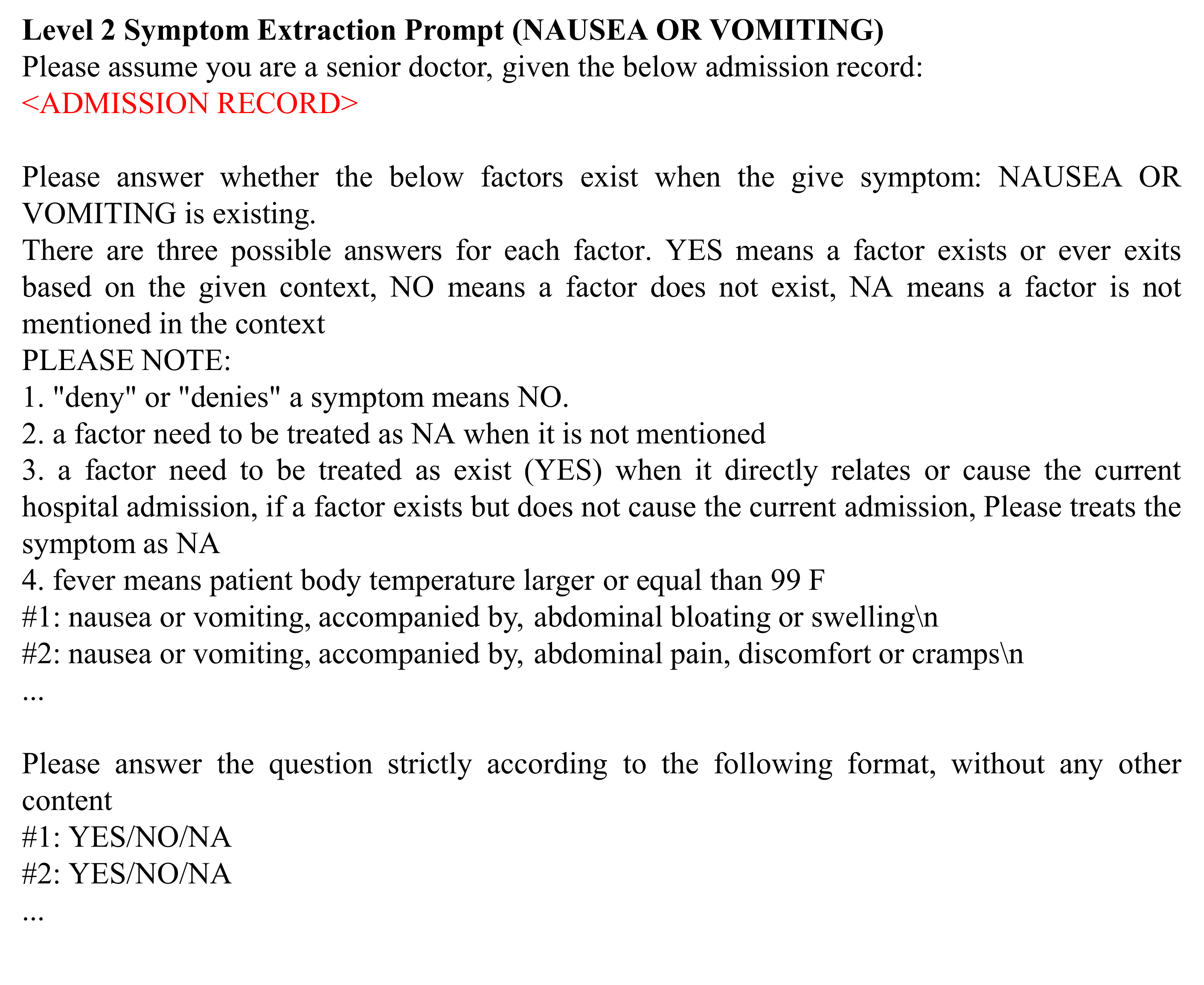}
  \caption{Admission Record Structurize Prompt 2}
  \label{fig:admissionStructurize2}
\end{figure}
\clearpage

\subsection{Patient Simulator Prompt} \label{sec:patientSimulator}
\begin{figure}[htbp]
  \centering
  \includegraphics[width=13.9cm,height=6.24cm]{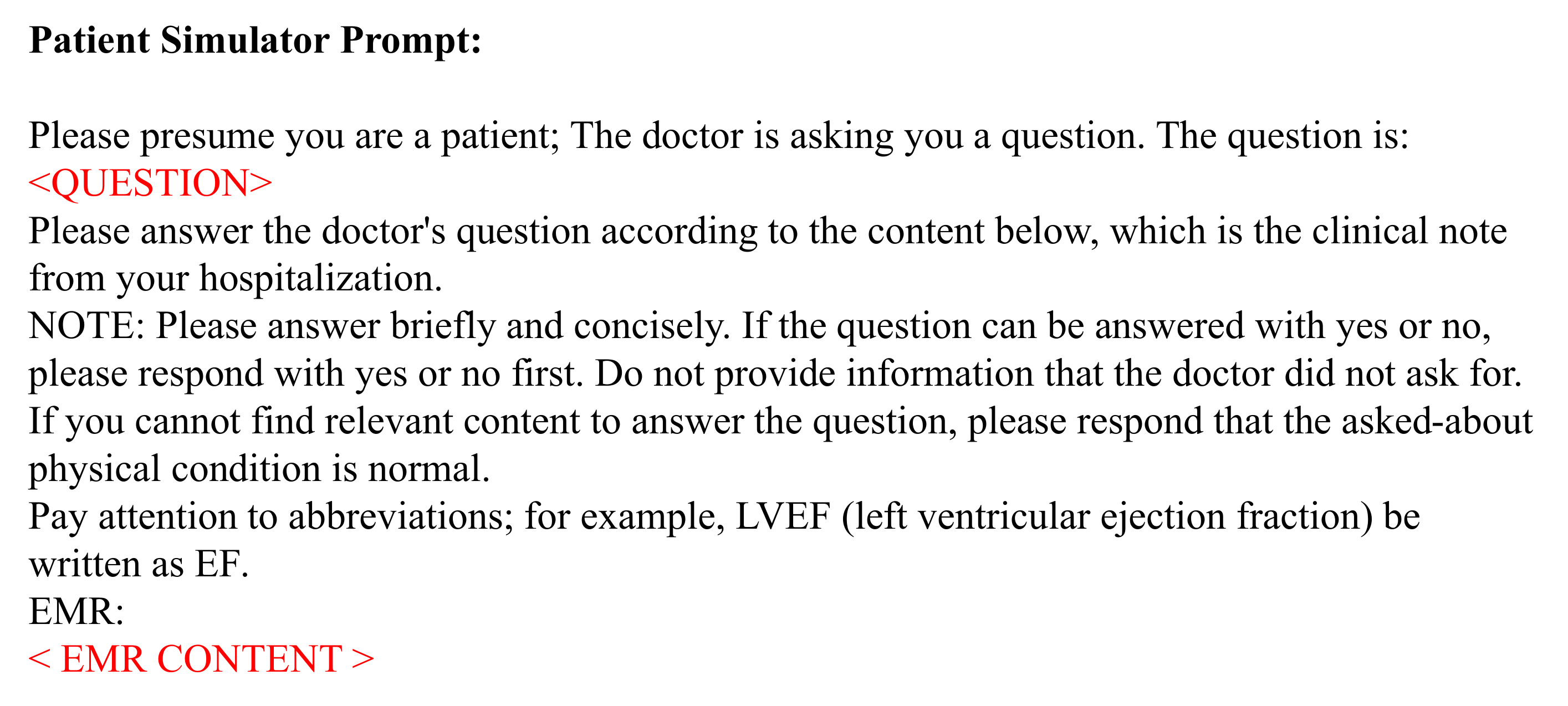}
  \caption{Patient Simulator Prompt}
  \label{fig:patientSimulator}
\end{figure}
\clearpage

\subsection{Screen Doctor Simulator Prompt} \label{sec:screeningDoctorSimulator}
\begin{figure}[htbp]
  \centering
  \includegraphics[width=13.9cm,height=8.38cm]{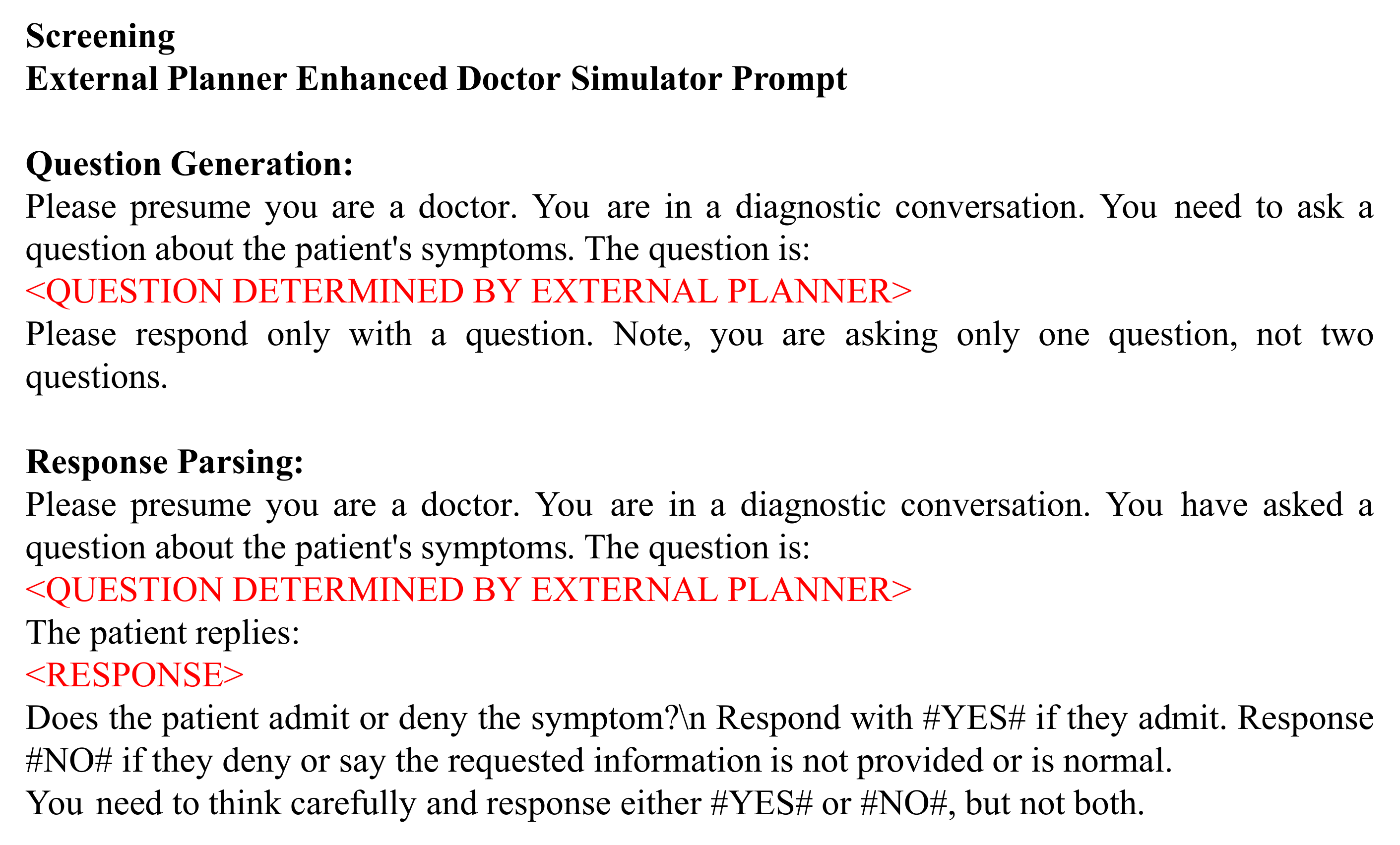}
  \caption{External Planner Enhanced Doctor Simulator Prompt (Screen)}
  \label{fig:screenDoctorSimulatorEP}
\end{figure}

\begin{figure}[htbp]
  \centering
  \includegraphics[width=13.9cm,height=13.94cm]{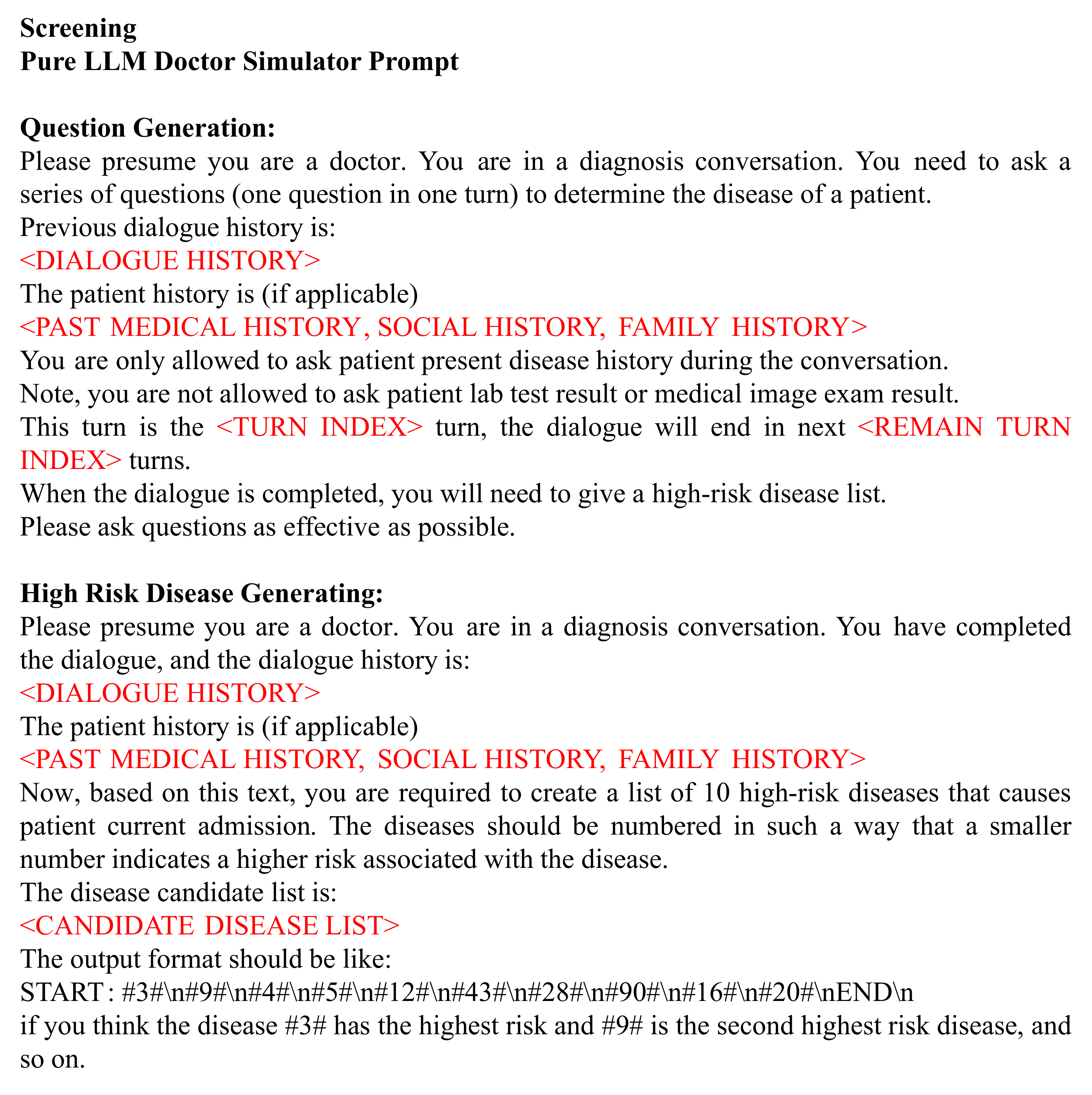}
  \caption{Pure LLM Doctor Simulator Prompt (Screen)}
  \label{fig:screenDoctorSimulatorPureLLM}
\end{figure}

\clearpage

\subsection{Differential Diagnosis Doctor Prompts} \label{sec:diagnosisDoctorSimulator}
\begin{figure}[htbp]
  \centering
  \includegraphics[width=13.9cm,height=7.1cm]{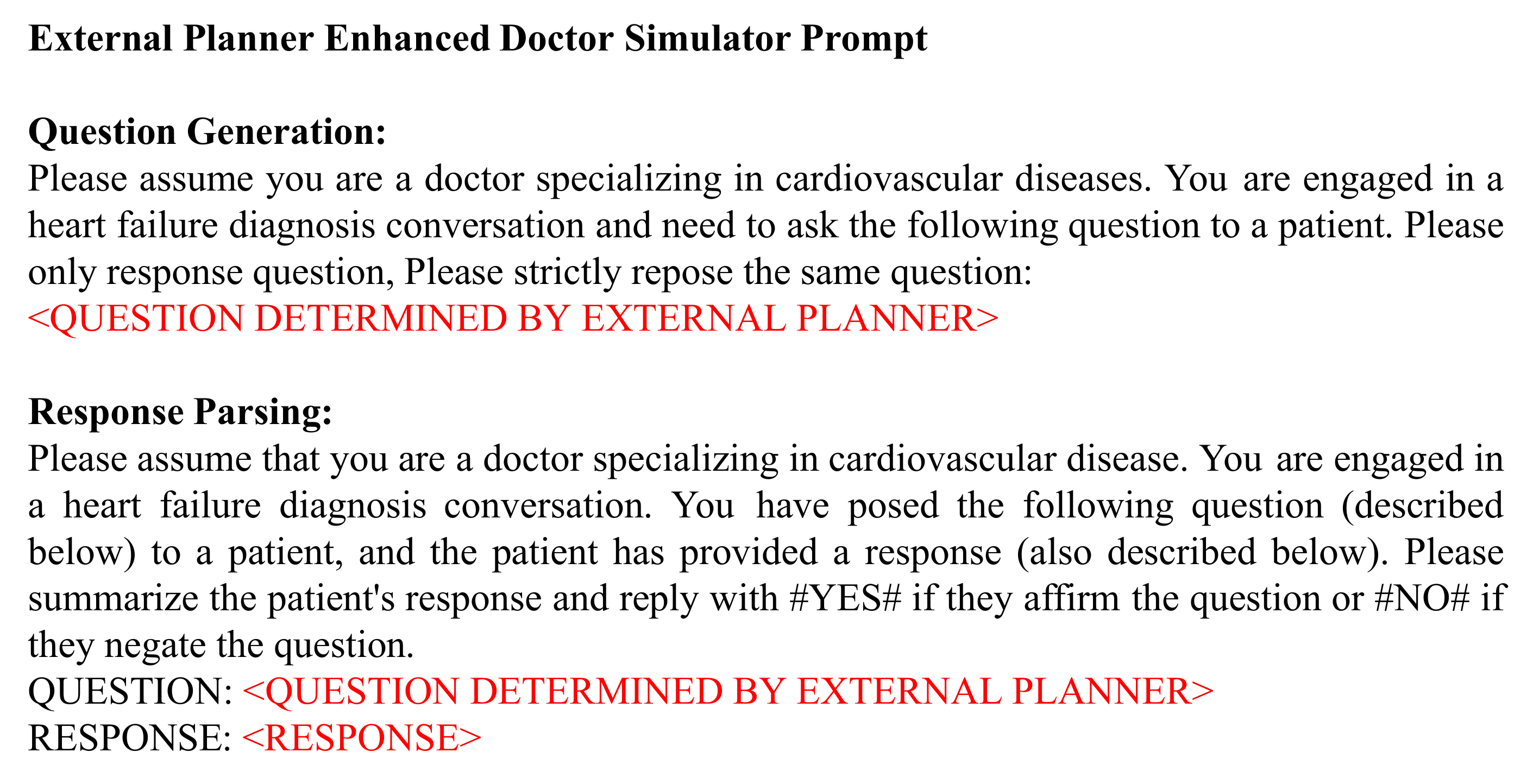}
  \caption{External Planner Enhanced Doctor Simulator Prompt (Diagnosis)}
  \label{fig:DiagnosisDoctorSimulatorEP}
\end{figure}

\begin{figure}[htbp]
  \centering
  \includegraphics[width=13.9cm,height=5.81cm]{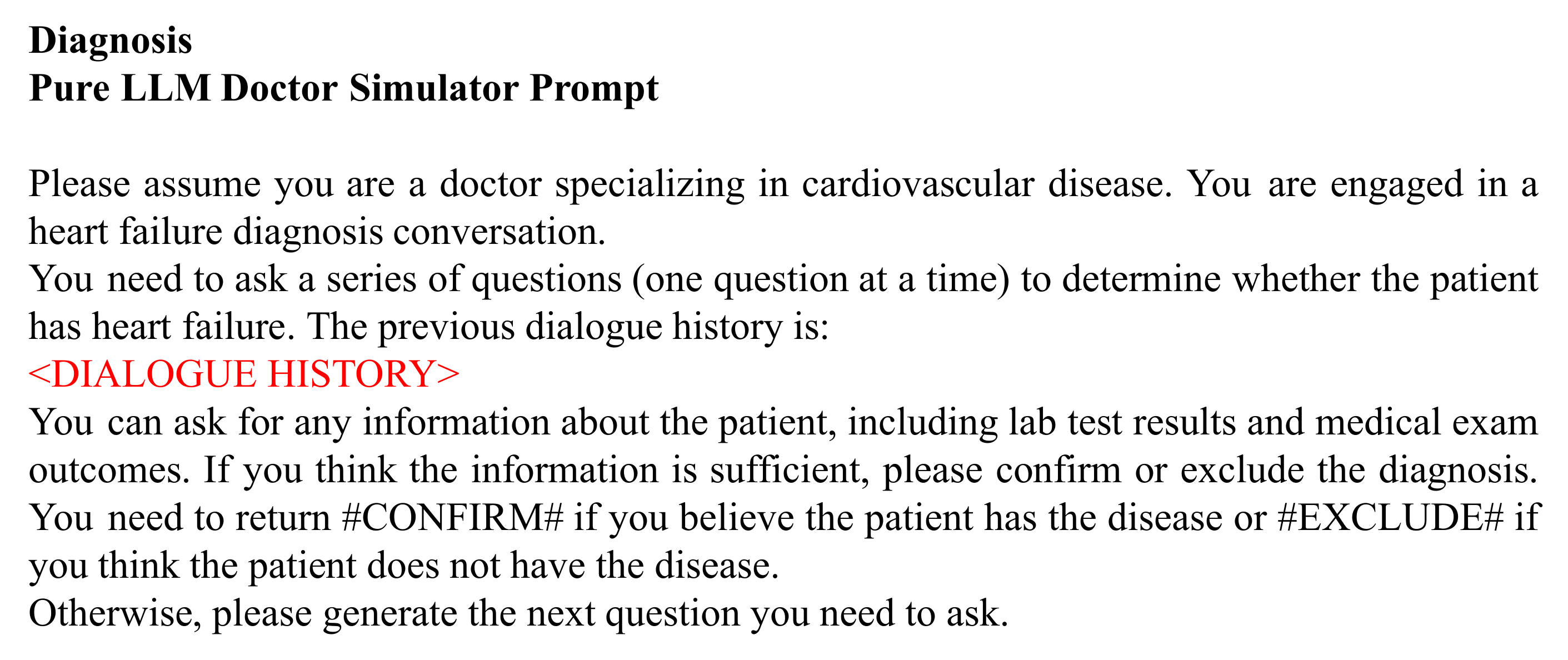}
  \caption{Pure LLM Doctor Simulator Prompt  (Diagnosis)}
  \label{fig:diagnosisDoctorSimulatorPureLLM}
\end{figure}

\begin{figure}[htbp]
  \centering
  \includegraphics[width=13.9cm,height=7.1cm]{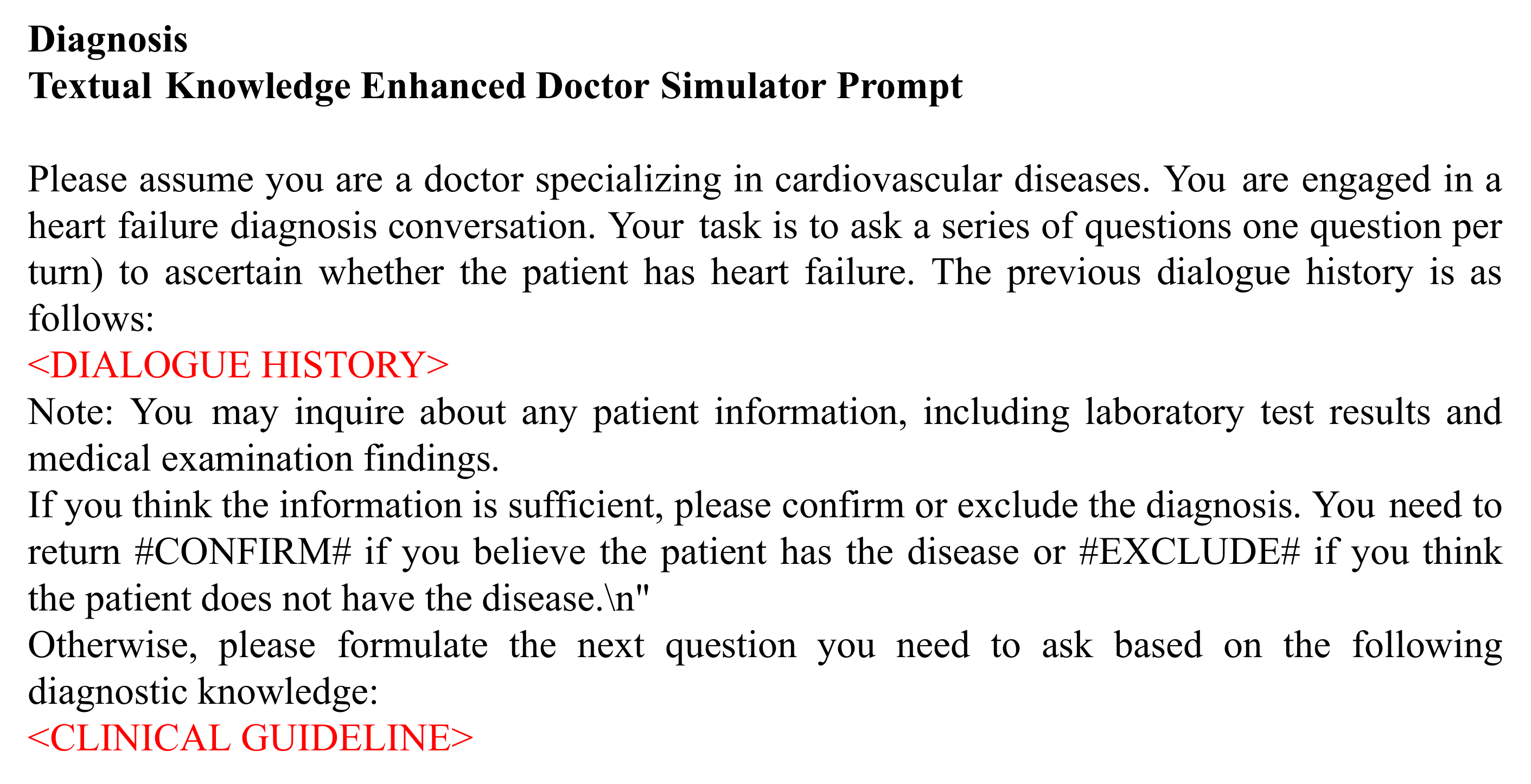}
  \caption{Textual Knowledge Enhanced Doctor Simulator Prompt (Diagnosis)}
  \label{fig:diagnosisDoctorSimulatorTK}
\end{figure}
\clearpage

\subsection{Decision Procedure Generator} \label{sec:decisionProceduregenerator}
\begin{figure}[htbp]
  \centering
  \includegraphics[width=13.9cm,height=16.07cm]{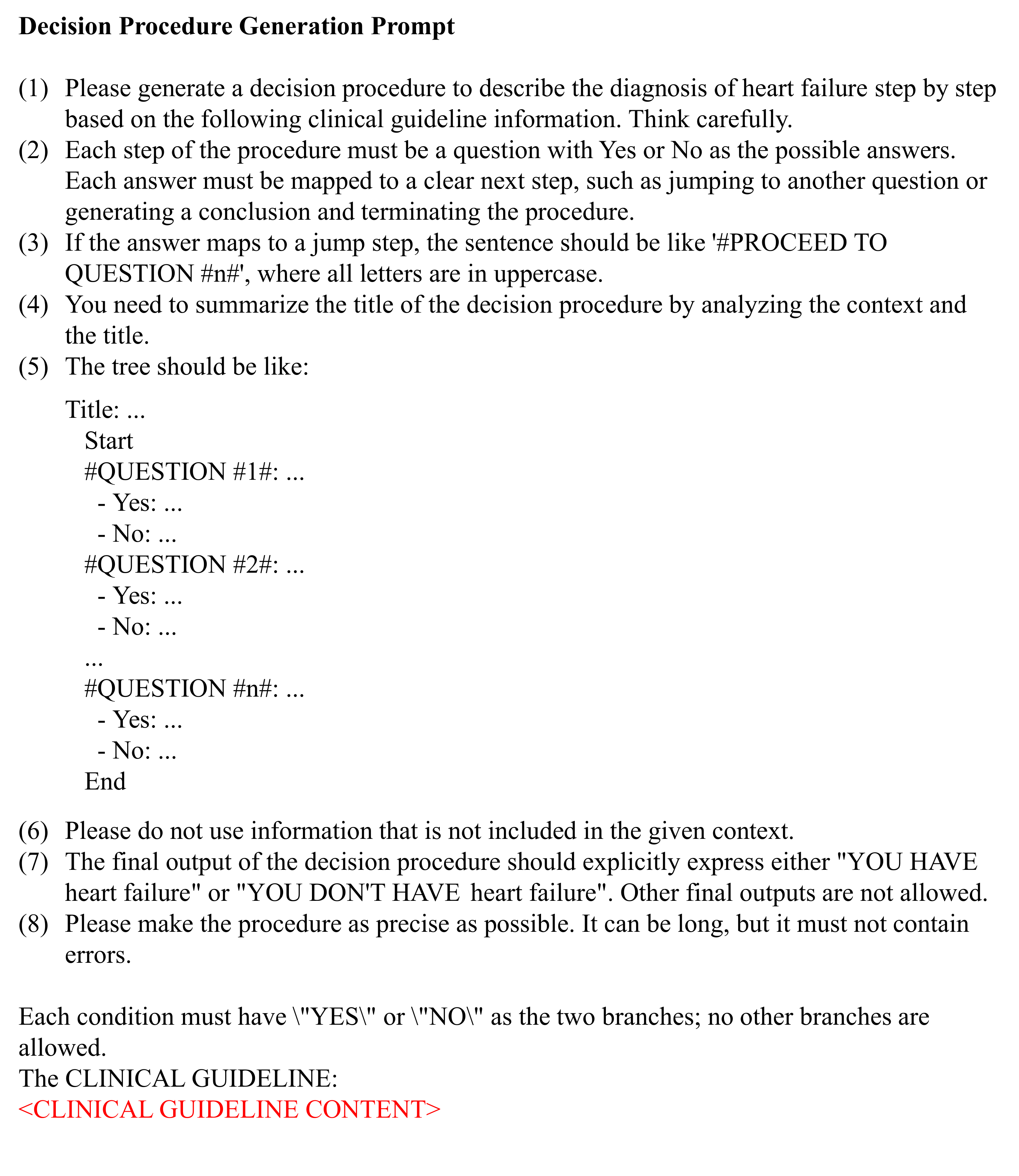}
  \caption{Decision Procedure Generate Prompt}
  \label{fig:decisionProcedureGenerator}
\end{figure}
\clearpage

\subsection{Dialogue Review} \label{sec:dialogueReview}
\begin{figure}[htbp]
  \centering
  \includegraphics[width=13.79cm,height=8.81cm]{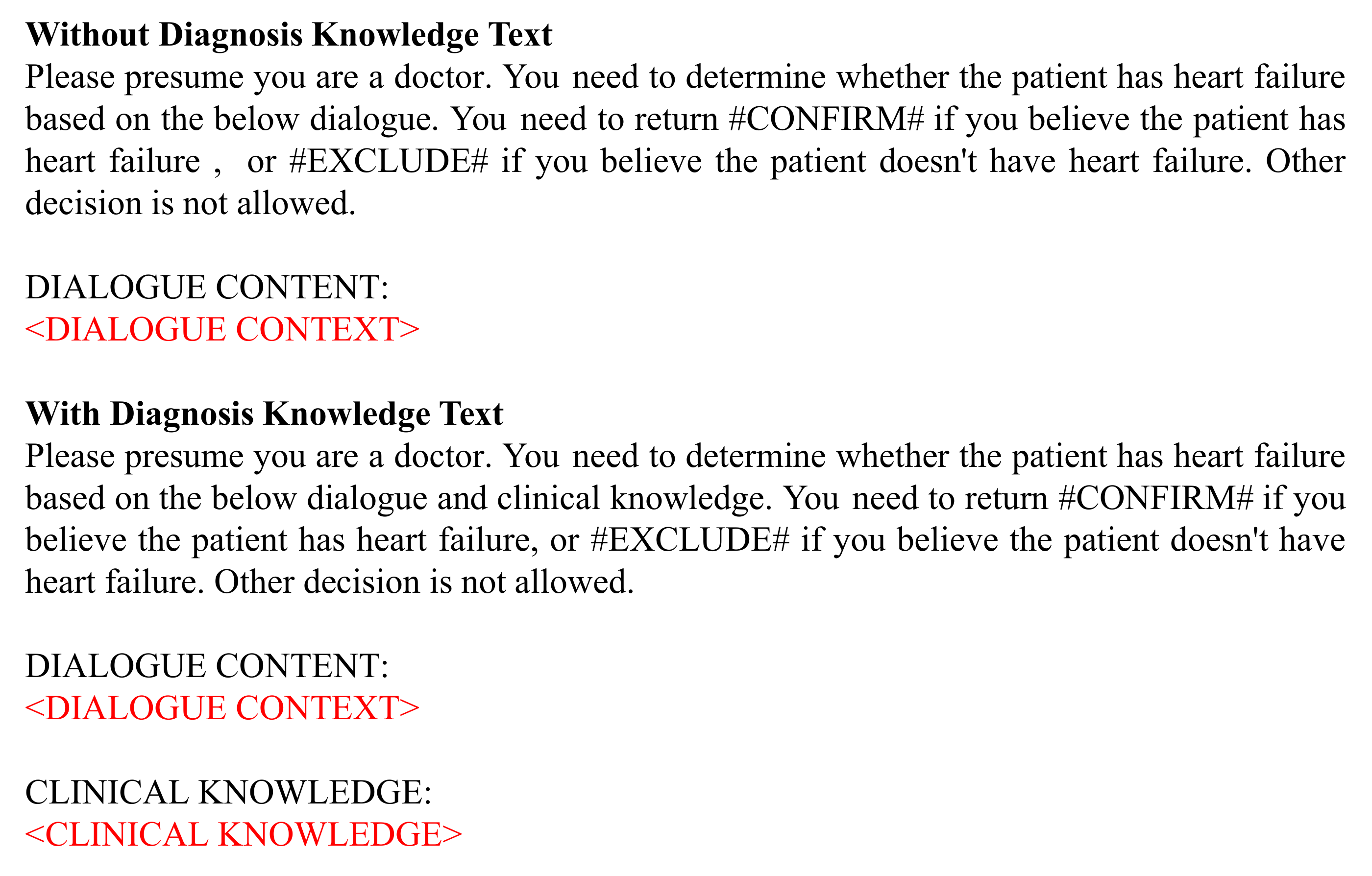}
  \caption{Dialogue Review Prompt}
  \label{fig:dialogueReview}
\end{figure}
\clearpage

\section{Patient EMR Sample} \label{sec:emrSample}
\begin{figure}[htbp]
  \centering
  \includegraphics[width=13.9cm,height=16.5cm]{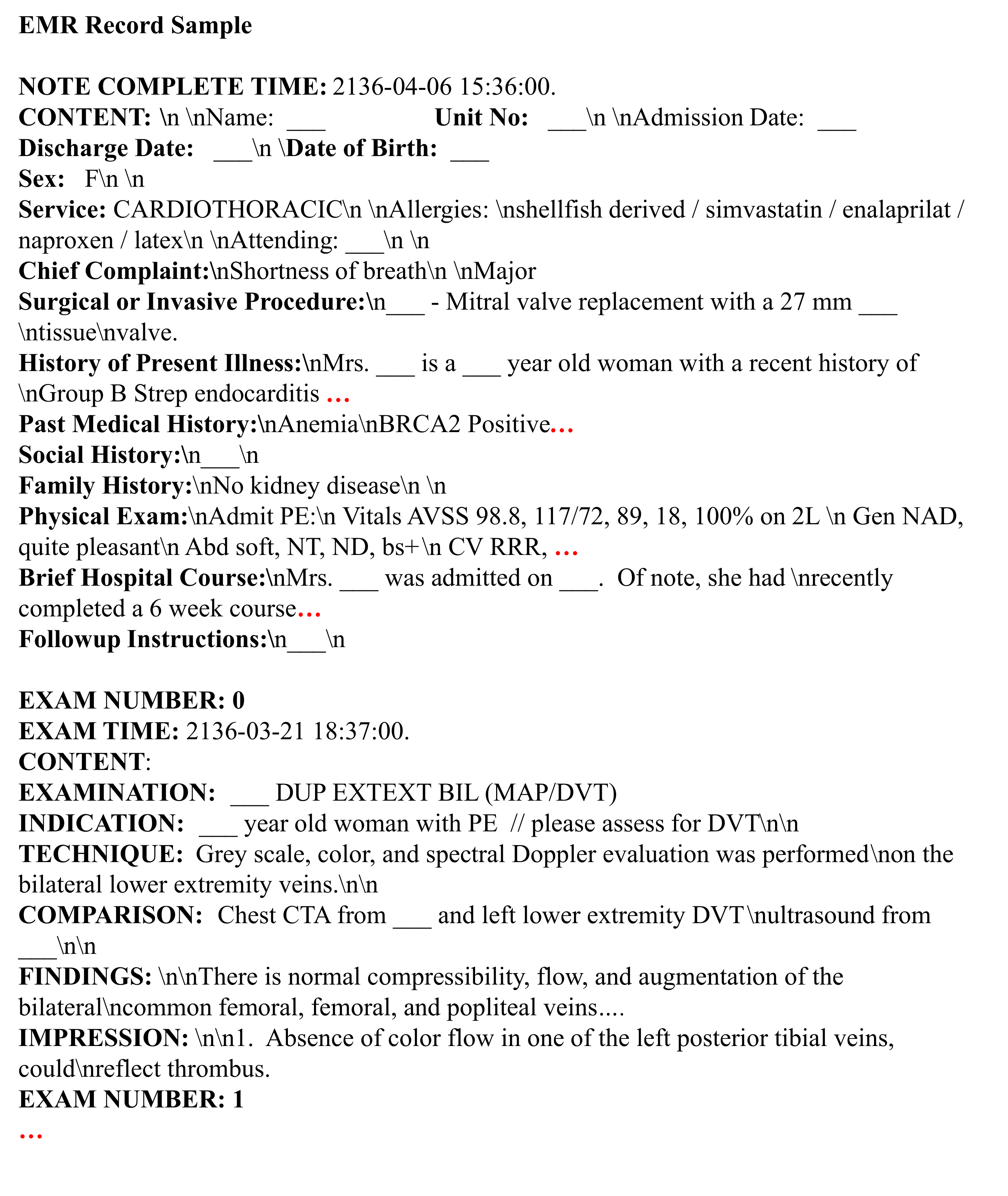}
  \caption{Patient EMR Sample}
  \label{fig:patientEMRSample}
\end{figure}
\clearpage

\section{Selected Screening Disease List} \label{sec:diseaseList}
\begin{figure}[h]
  \centering
  \includegraphics[width=13.9cm,height=17.7cm]{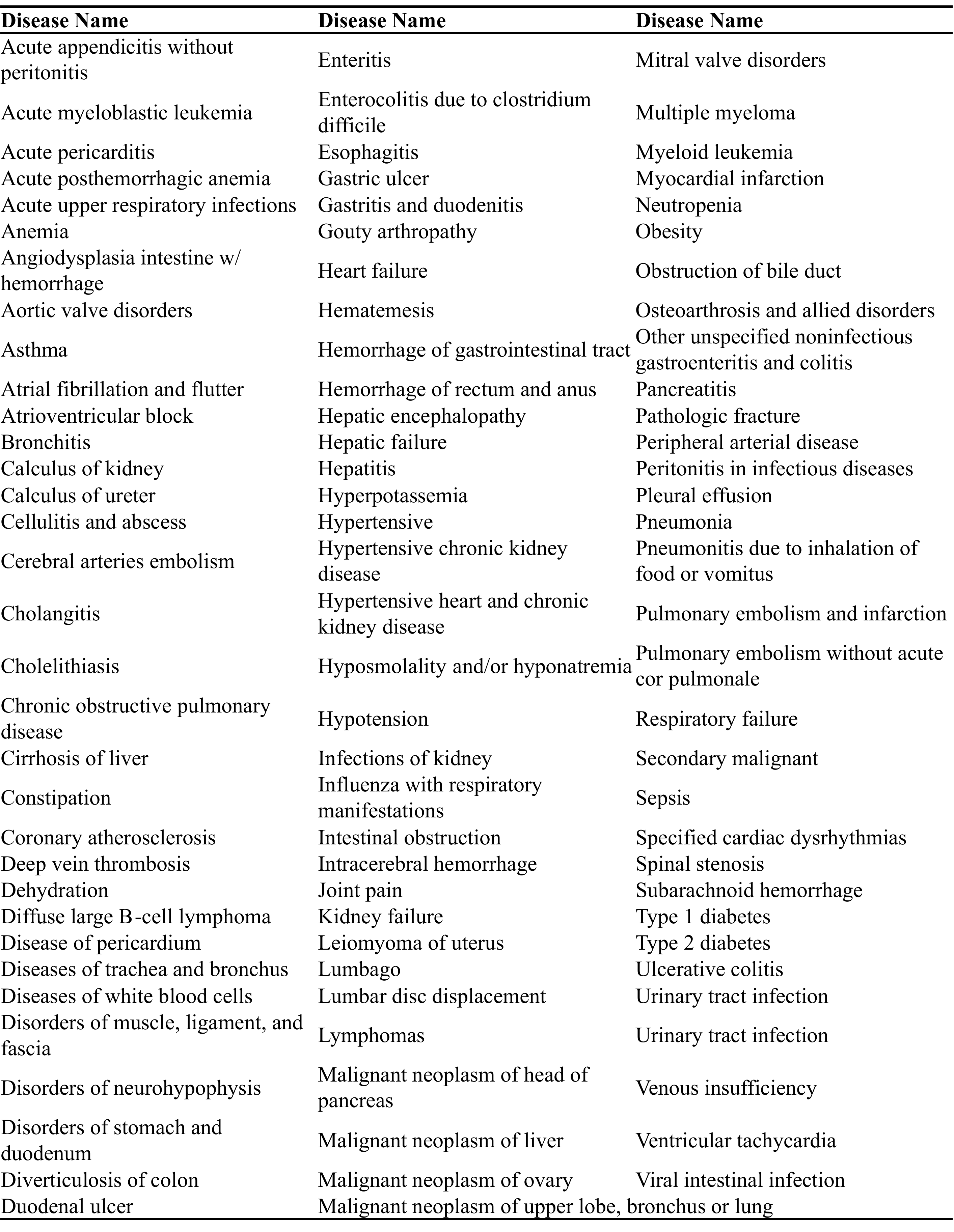}
  \caption{Selected Diseases List}
  \label{fig:diseaseList}
\end{figure}

\end{appendices}

\end{document}